%% file: main.tex
\pdfoutput=1

\documentclass[11pt]{article}

\usepackage[preprint]{acl}

\usepackage{times}
\usepackage{latexsym}

\usepackage[T1]{fontenc}

\usepackage[utf8]{inputenc}

\usepackage{microtype}

\usepackage{inconsolata}

\usepackage{graphicx}

\usepackage{inconsolata}
\usepackage{makecell}
\usepackage{enumitem}
\usepackage{xspace}
\usepackage{graphicx}
\usepackage{multirow}
\usepackage{multicol}
\usepackage{booktabs}
\usepackage[ruled]{algorithm2e}
\usepackage{amssymb}
\usepackage{colortbl}
\usepackage[table,xcdraw]{xcolor}
\usepackage{amsmath}
\usepackage{subcaption}
\definecolor{mycolor}{HTML}{4285F4}
\newcommand{\name}{LLaMAX2\xspace}
\newcommand{\method}{layer-selective tuning\xspace}
\newcommand{\modelsmall}{Qwen3-XPlus-8B\xspace}
\newcommand{\modellarge}{Qwen3-XPlus-14B\xspace}
\newcommand{\modelseries}{Qwen3-XPlus\xspace}

\title{\name: Your Translation-Enhanced Model also Performs Well in Reasoning}

\newcommand*{\affaddr}[1]{#1} 
\newcommand*{\affmark}[1][*]{\textsuperscript{#1}}
\newcommand*{\email}[1]{\texttt{#1}}
\author{
\textbf{Changjiang Gao}\affmark[$\heartsuit$]\thanks{Work done during internship at Shanghai Artificial Intelligence Laboratory}, \textbf{Zixian Huang}\affmark[$\clubsuit$], \textbf{Jingyang Gong}\affmark[$\diamondsuit\clubsuit$], \\
\textbf{Shujian Huang}\affmark[$\heartsuit$], 
\textbf{Lei Li}\affmark[$\spadesuit$], \textbf{Fei Yuan}\affmark[$\clubsuit$]\\
\affaddr{\affmark[$\heartsuit$]National Key Laboratory for Novel Software Technology, Nanjing University} \\
\affaddr{\affmark[$\diamondsuit$]The University of Hong Kong},
\affaddr{\affmark[$\spadesuit$]Carnegie Mellon University}, \\
\affaddr{\affmark[$\clubsuit$]Shanghai Artificial Intelligent Laboratory}\\
\email{gaocj@smail.nju.edu.cn}, \email{huangsj@nju.edu.cn}, \email{leili@cs.cmu.edu} \\
\email{jygong@hku.hk}, ~\email{\{huangzixian, yuanfei\}@pjlab.org.cn}\\
}

\begin{document}
\maketitle

\input{ACL_2026/sections/0.abstract}

\input{ACL_2026/sections/1.introduction}

\input{ACL_2026/sections/2.related_work}

\input{ACL_2026/sections/3.algorithm}

\input{ACL_2026/sections/4.experiments}

\input{ACL_2026/sections/5.analysis}

\input{ACL_2026/sections/6.conclusion}


\input{main.bbl}
\clearpage
\newpage
\appendix

\input{ACL_2026/sections/7.appendix}

\end{document}

%% file: ACL_2026/sections/0.abstract.tex
\begin{abstract}


General Large Language Models (LLMs) excel in reasoning, but those enhanced for translation struggle with reasoning tasks. To address this, we propose a novel translation-enhanced recipe that begins with instruct models and applies layer-selective tuning only on parallel data. Following this pipeline, we introduce the \modelseries models, which demonstrate significant improvements in translation performance across both high- and low-resource languages, achieving 15+ spBLEU and 40+ xComet in low-resource languages, like Swahili. Interestingly, training only with small parallel datasets, \modelseries achieves an average improvement of 1+ points on 7 multilingual tasks while maintaining proficiency comparable to the Qwen3 instruct model in 15 popular reasoning datasets. This work offers a promising approach to multilingual enhancement, significantly reducing complexity and enhancing accessibility for a wider range of languages. The code~\footnote{\url{https://github.com/CONE-MT/LLaMAX2.0}} and model~\footnote{\url{https://huggingface.co/collections/LLaMAX/llamax20-68ad1c154fcf2623b75a068c}} are publicly available.

\end{abstract}

%% file: ACL_2026/sections/1.introduction.tex
\section{Introduction}

Large Language Models~(LLMs; \citealp{IntroducingGPT52025,IntroducingClaude4,comaniciGemini25Pushing2025,deepseek-aiDeepSeekR1IncentivizingReasoning2025a,teamKimiK2Open2025,ByteDanceSeed,yangQwen3TechnicalReport2025a}) excel in reasoning; however, translation-enhanced LLMs~\cite{alves2024tower,reiTowerBridgingGenerality2025, sunHunyuanLargeOpenSourceMoE2024, lu-etal-2024-llamax} face challenges in this area. As shown in Figure~\ref{fig:introduction_general}, Tower-Plus-9B~\cite{reiTowerBridgingGenerality2025}, a translation-enhanced model, significantly improves multilingual instruction-following capabilities, yet underperforms on reasoning, such as LiveCodeBench-V5~\cite{jainLiveCodeBenchHolisticContamination2024a} and AIME2025~\cite{ArtProblemSolving}.

The root of this dilemma lies in the limitations of the current training approach~\cite{reiTowerBridgingGenerality2025,lu-etal-2024-llamax,douSailor2SailingSouthEast2025,zhengHunyuanMTTechnicalReport2025}, which typically begins with a base model and then trains on large multilingual datasets. The preference for starting with a base model instead of an instruct model stems from the fact that full fine-tuning~(FFT) can result in catastrophic forgetting~\cite{li-etal-2024-revisiting,alexandrov-etal-2024-mitigating}.

\input{ACL_2026/images/introduction_general_capabilities}

\input{ACL_2026/images/introduction}

Unlike previous work, \modelseries are built on Qwen3 instruct models rather than base models. Since fundamental reasoning skills like math and coding are universal, there is no need to learn basic concepts in multiple languages~\cite{benchmax:2025,gao2025thinkingmultilinguallyempowerllm}. Meanwhile, to mitigate catastrophic forgetting, we apply \method which effectively balances translation quality and reasoning capabilities without the need for extra parameters. It employs a two-phase tuning process, training the four layers closest to the embedding layer and the fifteen layers further away, which consistently yields significant improvements across various datasets and model backbones. As a result, \modelseries significantly reduce the reliance on large amounts of high-quality data.

Particularly, a small amount of parallel data is enough for translation-enhancment. As depicted in Figure~\ref{fig:introduction_main}~\footnote{x includes Spanish, French, German, Russian, Bengali, Japanese, Thai, Swahili, Chinese, Telugu, Arabic, Korean, Serbian, Czech, Hungarian, and Vietnamese}, Tower-Plus-9B has an impressive 32 billion tokens, LLaMAX~\cite{lu-etal-2024-llamax} features 60 billion tokens, Sailor2-8B-Chat~\cite{douSailor2SailingSouthEast2025} needs 500 billion tokens, and Hunyuan-MT~\cite{zhengHunyuanMTTechnicalReport2025} takes it even further with 1311 billion tokens. 
Not to mention, higher-quality data are needed for the supervised fine-tuning~\citealp{chungScalingInstructionFinetunedLanguage2024a,ouyangTrainingLanguageModelsFollow2022}), which poses a significant challenge for low-resource languages. Howerver, \modelseries utilize only 0.8 billion tokens from with our careful pre-processing. We standardize the original data from NLLB~\cite{nllbteam2022languageleftbehindscaling,schwenk-etal-2021-ccmatrix} and OPUS-100~\cite{tiedmann2012parallel} to a format-unified, clean, deduplicated, language-consistent, quality-controlled, and instructed-formatted dataset. 

\modelseries significantly improves translation performance, achieving over 15+ spBLEU points increase and 40+ xComet points in low-resource~(sw), with notable enhancements in high-resource translations. Utilizing small parallel data alone, it also demonstrates an average improvement of over 1 point across 7 multilingual tasks, including xIFEval~\cite{benchmax:2025}, MGSM~\cite{shiLanguageModelsAre2022b}, XGPQA~\cite{rein2024gpqa, benchmax:2025}, and so on. Furthermore, comprehensive testing on 15 popular reasoning datasets, such as BBEH~\cite{kazemi-etal-2025-big}, Livecodebench~\cite{jainLiveCodeBenchHolisticContamination2024a}, Olymmath~\cite{he-etal-2024-olympiadbench} and so on, shows that it surpasses existing translation-enhanced models and performs on par with Qwen3 instruct models. Our main contributions can be summarized as follows:
\begin{itemize}
[nosep,itemsep=2pt,leftmargin=0.3cm]
    \item We propose a new training recipe: using a small amount of parallel data for layer-selective tuning on an instruct model, which significantly reduces complexity and makes it more accessible for a wider range of languages.
    
    \item We introduce 2 open-sourced translation-enhanced \modelsmall and \modellarge, which maintaining reasoning capabilities.
    \item Extensive experiments on \modelseries and comprehensive benchmark evaluations demonstrate that we can achieve a balance between translation quality and reasoning capabilities.
\end{itemize}

%% file: ACL_2026/images/introduction_general_capabilities.tex
\begin{figure}[!t]
    \centering
    \includegraphics[width=\linewidth]{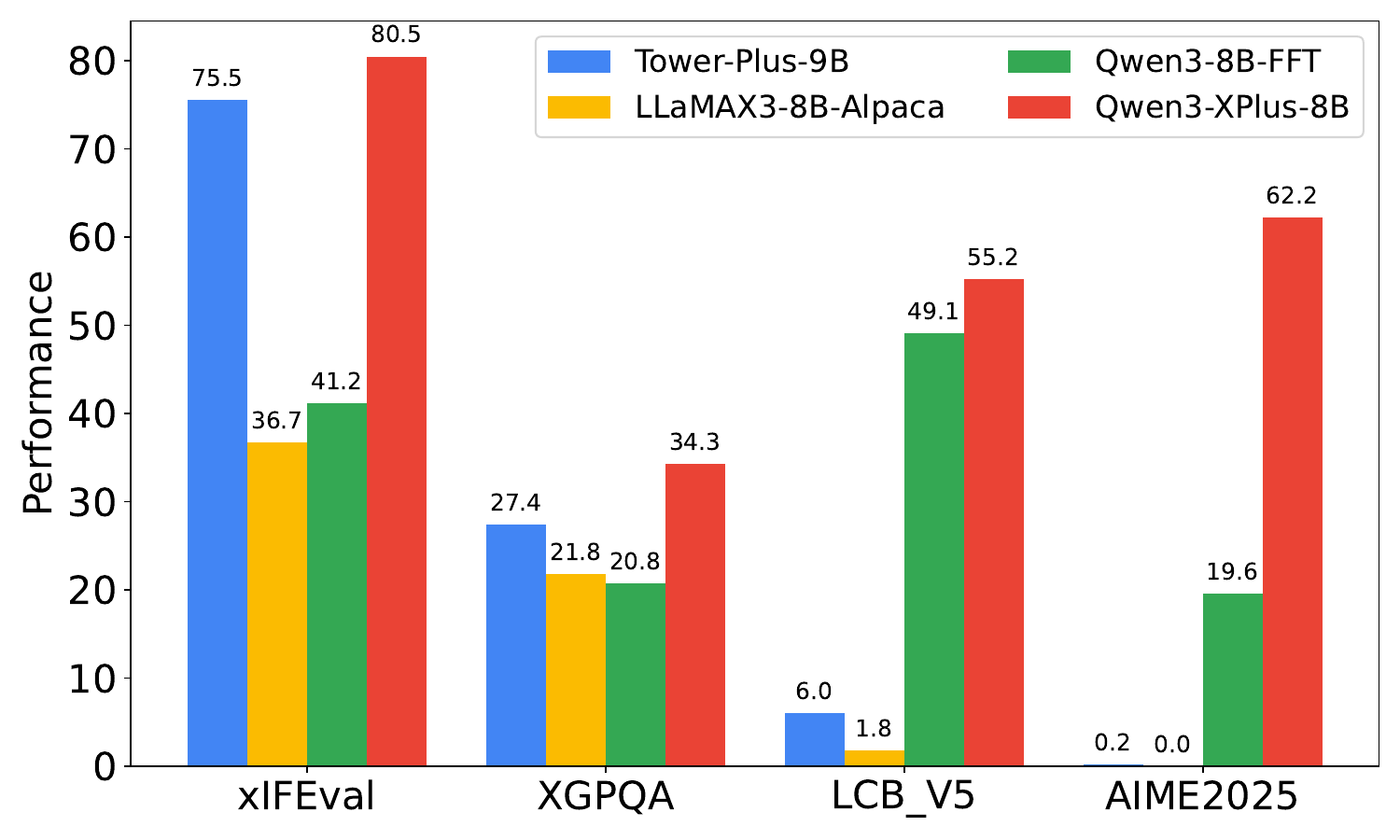}
    \caption{Comparison of translation-enhanced models on reasoning tasks. Tower-Plus-9B and LLaMAX-3-Alpaca struggle with LiveCodeBench-V5~(LCB\_V5) and AIME2025, whereas \modelsmall effectively addresses these challenges.}
    \label{fig:introduction_general}
\end{figure}

%% file: ACL_2026/images/introduction.tex
\begin{figure*}[!h]
    \centering
    \includegraphics[width=\textwidth]{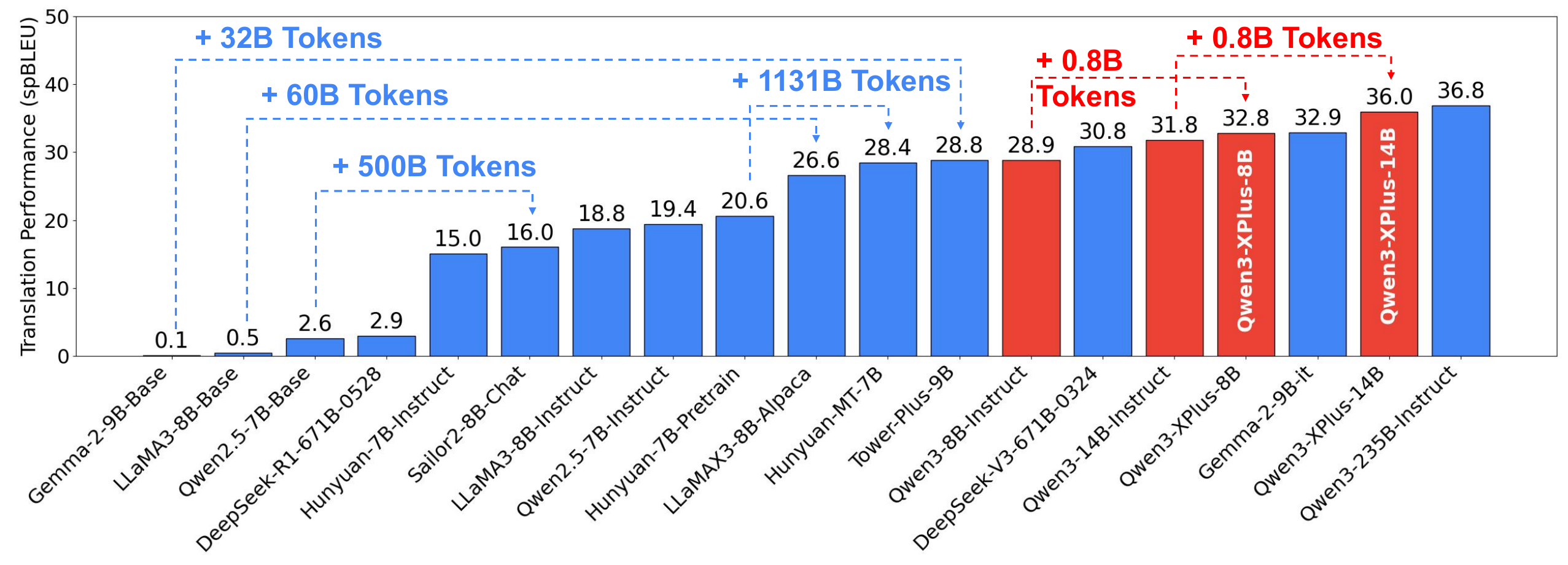}
    \caption{Average translation performance from English to 16 languages~(en$\rightarrow$x). Unlike previous methods that train from a base model, \modelseries begins with an instruct model and, using limited parallel data, achieves significant improvements in translation.}
    \label{fig:introduction_main}
\end{figure*}

%% file: ACL_2026/sections/2.related_work.tex
\section{Related Work}
\subsection{Massively Multilingual Translation with Large Language Models}
Massively multilingual translation refers to building a single machine translation model to handle many language directions \cite{zhang-etal-2020-improving}. 
Due to the multilingual and instruction-following nature of LLMs, they already show high translation performance in many directions without training~\cite{bawden-yvon-2023-investigating,zhu-etal-2024-multilingual} or with minimal training \cite{li-etal-2024-eliciting,cui-etal-2025-multilingual,MindMerger}. Based on this, specialized translation LLMs have been developed to do massive multilingual translation. For example, LLaMAX \cite{lu-etal-2024-llamax} and Tower \cite{alvesTowerOpenMultilingual2024a} apply continued pretraining and instruction tuning on the LLaMA-2 model \cite{touvron2023llama2openfoundation} with massive parallel and monolingual data, achieving comparable performance to specialized translation models. MT-R1-Zero \cite{feng2025mtr1zeroadvancingllmbasedmachine} adapts the R1-Zero reinforcement learning framework to the translation task and resulting a reasoning translation model. However, the models' general instruction-following and reasoning capabilities drop after the training, which weakens the advantage of using LLMs for translation. This work finds a simple, parameter-efficient but effective approach to train a translation LLM while keeping (and even improving) its general ability.

\subsection{Parameter-Efficient Finetuning}
There are many parameter-efficient finetuning~(PEFT) techniques for tuning LLMs with less resource and reduced catastrophic forgetting. According to \citeposs{hanParameterEfficientFineTuningLarge2024} survey, there are mainly four types of PEFT: (1) Additive, including adapters \cite{houlsbyParameterEfficientTransferLearning2019a,pfeiffer-etal-2021-adapterfusion,heUnifiedViewParameterEfficient2021} and soft prompts \cite{liPrefixTuningOptimizingContinuous2021,liu-etal-2022-p}; (2) Selective \cite{guo-etal-2021-parameter,sungTrainingNeuralNetworks2021,liao-etal-2023-parameter}; (3) Reparameterized, mainly the LoRA family \cite{huLoRALowRankAdaptation2021b,dettmersQLoRAEfficientFinetuning2023,liuDoRAWeightDecomposedLowRank2024,owodunni2025continually}; (4) Hybrid \cite{heUnifiedViewParameterEfficient2021,hu-etal-2023-llm}. Our proposed method belongs to selective PEFT.

%% file: ACL_2026/sections/3.algorithm.tex
\section{\modelseries Training Recipe}

\input{ACL_2026/images/overview}

In this section, we outline our new translation enhancement recipe~(Figure \ref{fig:overview}). We train an instruct model with instruction-formatted parallel data~($\S$~\ref{sec:data_construction}) using layer-selective tuning~($\S$~\ref{sec:layer_selective_tuning}).

\subsection{Training Data Construction}
\label{sec:data_construction}
The parallel data used in our training process comes from two public accessible datasets, NLLB \cite{nllbteam2022languageleftbehindscaling} and OPUS-100 \cite{tiedmann2012parallel}, with 6 processing steps:
\begin{enumerate}
[nosep,itemsep=2pt,leftmargin=0.3cm]
    \item Transform the data into a unified JSONL format, containing \texttt{src, trg, src\_line, tgt\_line};
    \item Clean invalid characters and punctuations to avoid encoding and tokenization issues;
    \item Deduplicate the data in each language pair via SimHash \cite{manku2007detecting} based on language-specific tokenization and source-target length match. We first split the source and target texts into words or characters based on their languages, and then filter out the samples where either the source or the target text is too short (fewer than 2 tokens) or length mismatch (one's length is less than 0.3 of the other's). Then, we concatenate the source and target of each sample, and calculate the SimHash conflicts between samples, deleting those with more than 2 conflicts;
    \item Filter out the samples with incorrect language labels identified by fasttext \footnote{\url{https://huggingface.co/facebook/fasttext-language-identification}};
    \item Evaluate the translation quality of each sample with the conditional loss of a small translation model (NLLB-200-Distilled-600M \footnote{\url{https://huggingface.co/facebook/nllb-200-distilled-600M}}), ruling out the samples with higher loss than 90\% of the FLORES-101 development samples.
    \item  Convert the data into instruction format by adding clear and diverse task instructions.
\end{enumerate}

\input{ACL_2026/images/single_lyr_tuning}

\subsection{Layer-Selective Tuning}
\label{sec:layer_selective_tuning}

\paragraph{Behavioral importance of the middle layers.} To investigate how training different layers affects model behavior to guide the choice of target layers, we conduct experiments where each layer is independently trained. The results, illustrated in Figure~\ref{fig:single_lyr_tuning}, highlight distinct behaviors across layers, revealing that training on intermediate layers, especially layer 20, results in a significant decline in translation performance. This finding highlights the significance of middle layers~\cite{uncovering_middle_repre,exploring_intermediate_layers}, which have been demonstrated to capture more general and transferable representations.

\paragraph{Gradient-based sensitivity results guide layer selection.} In addition to single-layer training, we analyze the nuclear norm, which measures the magnitude of the gradient~\cite{liHowInstructionReasoning2025}, reflecting the sensitivity of model parameters to changes in input, thus revealing the stability and robustness of layers during training. 
Figure \ref{fig:nuclear_norm} illustrates the layer-wise nuclear norm~(the parameter sensitivity of $Q$, $K$, and $V$ matrices) for ``en-zh'' data in the FLORES-101 development dataset, with similar trends observed in other translation directions.

\paragraph{Internal encoder-decoder hypothesis inspires a two-stage training approach.} In decoder-only models, bottom layers~(close to the input embedding layer) primarily focus on encoding information, while top layers~(far away from the input embedding layer) emphasize the decoding process. By leveraging this insight, as discussed in previous studies~\cite{chen2024image,lin2025boostingmultimodallargelanguage}, we enhance training strategies to optimize the performance of decoder-only models. 

\input{ACL_2026/images/nuclear_norm}

\subsection{Training Algorithm}
\input{ACL_2026/tabs/training_recipe}

As shown in Algorithm~\ref{alg:training_recipe}, we fine-tune an instruct model using the same instruction-formatted parallel data in two stages. In Stage 1, we focus on tuning the bottom $k$ layers of the model. Then, in Stage 2, we focus on adjusting the top $m$ layers of the model that have already undergone training in the first stage. Throughout the tuning process, the parameters of the middle layers are frozen.

%% file: ACL_2026/images/overview.tex
\begin{figure}[!t]
    \centering
    \includegraphics[width=\linewidth]{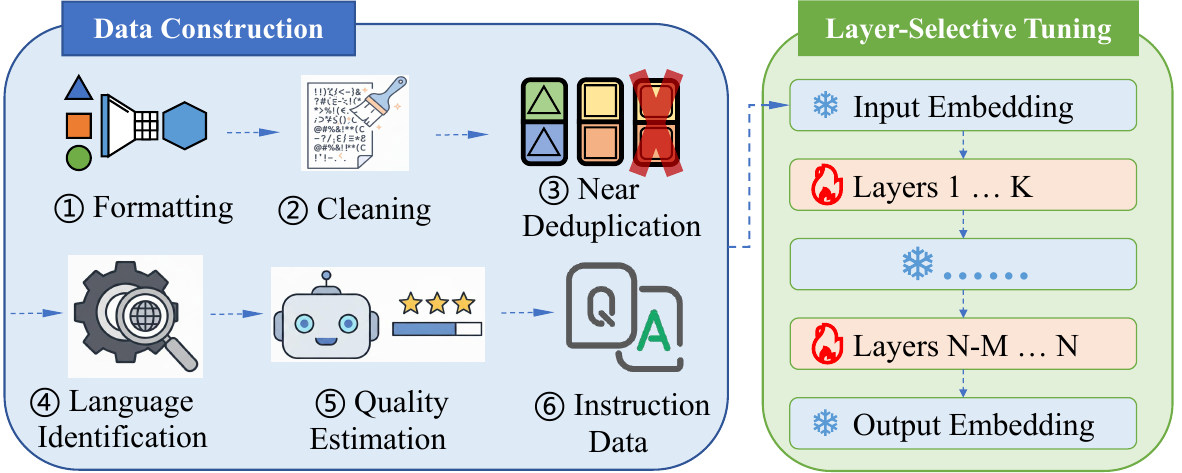}
    \caption{Overview of \modelseries training recipe. After the data construction process, an instruct model is trained using layer-selective tuning strategy with instruction-format parallel data.}
    \label{fig:overview}
\end{figure}

%% file: ACL_2026/images/single_lyr_tuning.tex
\begin{figure}[!t]
    \centering
    \includegraphics[width=\linewidth]{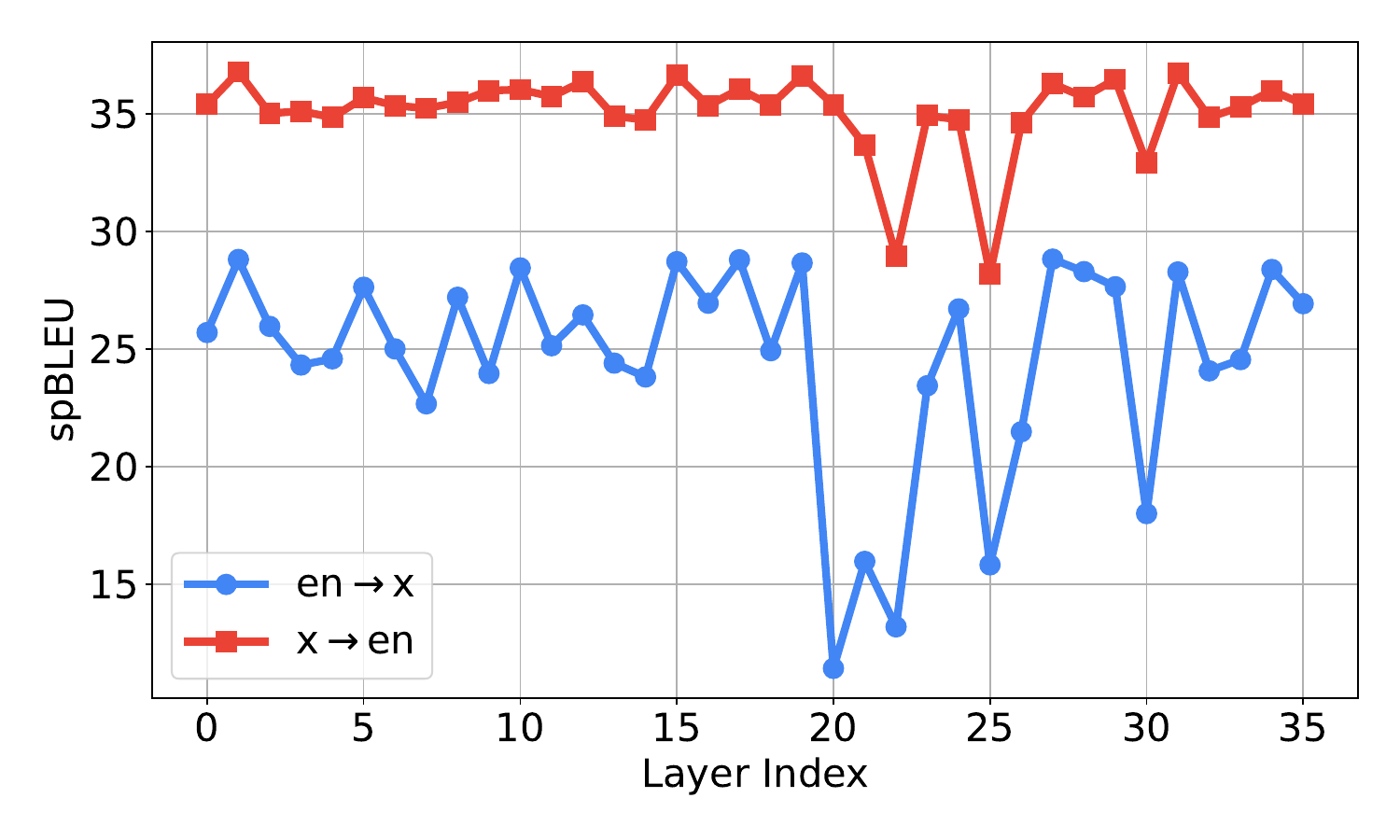}
    \caption{Translation performance of models that are single-layer tuned on parallel data.}
    \label{fig:single_lyr_tuning}
\end{figure}

%% file: ACL_2026/images/nuclear_norm.tex
\begin{figure}[!t]
    \centering
    \includegraphics[width=\linewidth]{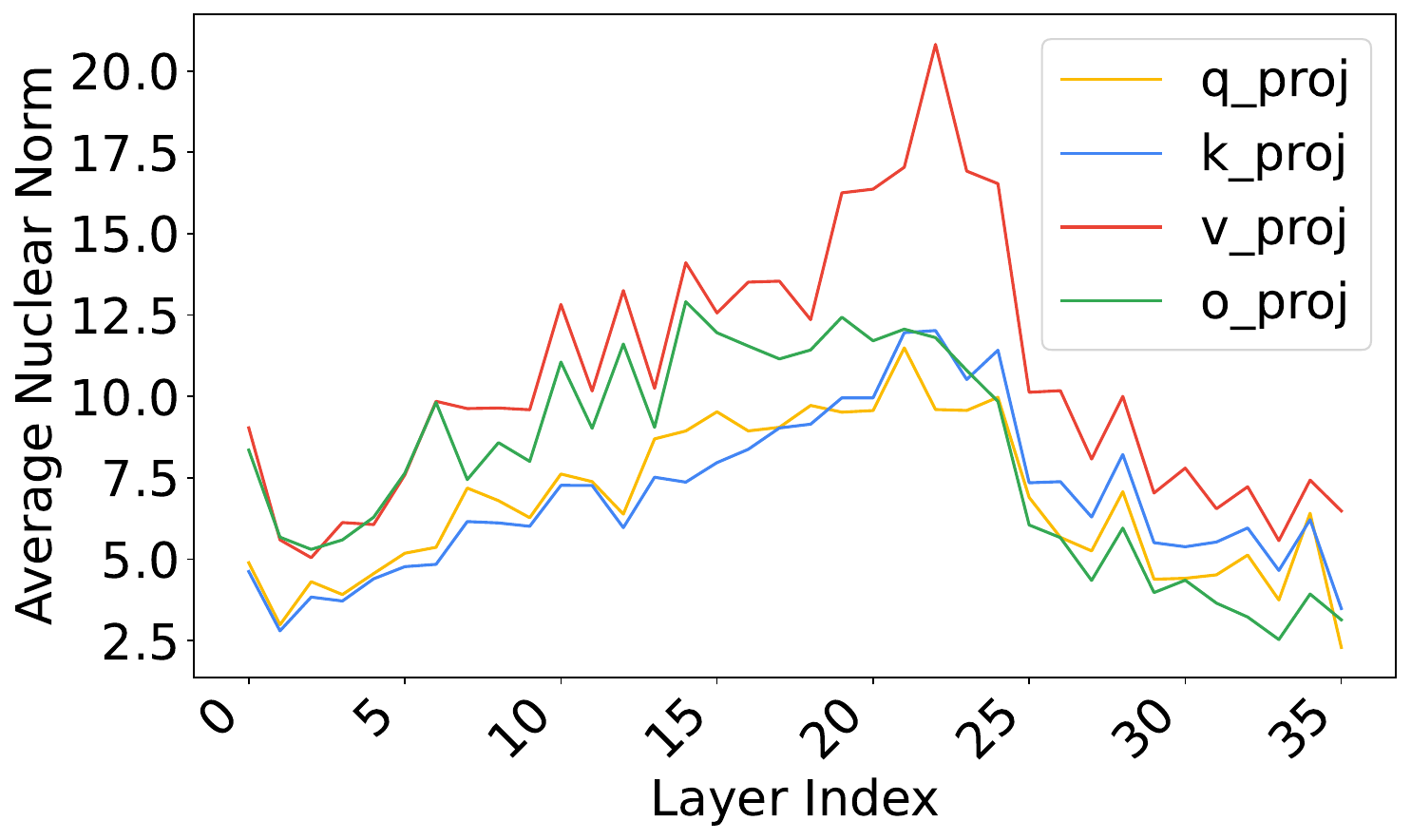}
    \caption{Layerwise nuclear norm of Qwen3-8B on the en-zh split of the Flores-101 dev dataset. About the 20th layers show the highest sensitivity in $Q$, $K$ and $V$.}
    \label{fig:nuclear_norm}
\end{figure}

%% file: ACL_2026/tabs/training_recipe.tex
\begin{algorithm}[t]
\footnotesize
\KwIn{Epoch number $L$, learning rate $\eta$. Training data for Stage 1 and Stage 2 is both $\mathcal{D}_{\mathrm{multi}}$, where $\mathcal{D}_{\mathrm{multi}} = \mathcal{D}_{\mathrm{en}\rightarrow \cdot} \bigcup \mathcal{D}_{\cdot \rightarrow \mathrm{en}}$. 
Given an instruct model, which pretrained parameter $\theta_0=\{ \theta_\mathrm{ie}, \theta_\mathrm{bottom\_k}, \cdots, \theta_\mathrm{top\_m}, \theta_\mathrm{oe}\}$, $\theta_\mathrm{ie}$ and $\theta_\mathrm{oe}$ are the embedding parameters, and $\theta_\mathrm{bottom\_k}, \cdots, \theta_\mathrm{top\_m}$ are parameters of transformer layers.The parameters of Stage 1 are initialized as $\theta_1=\theta_0$ with only $\theta_\mathrm{bottom\_k}$ being trainable, where $\theta_\mathrm{bottom\_k}=\{\theta_\mathrm{layer\_1}, \cdots,\theta_\mathrm{layer\_k} \}$. Note, the parameters used for Stage 2 are initialized as $\theta_2=\theta_1$, with only $\theta_\mathrm{top\_m}$ being trainable, where $\theta_\mathrm{top\_m} = \{ \theta_\mathrm{layer\_{n-m}}, \cdots, \theta_\mathrm{layer\_n} \}$. } 


\tcp{\color{mycolor} Stage 1, only $\theta_\mathrm{bottom\_k}$ being trainable.}

\For {epoch $l = 1$ to $L$}   
{
	Shuffle $\mathcal{D}_{\mathrm{multi}}$ to obtain a new training sequence. \\
	\For {each batch $\mathcal{D}_1 \in \mathcal{D}_{\mathrm{multi}} $} 
	{
        $l_1 = \sum_{\mathbf{x,y}\sim\mathcal{D}_1} - \mathrm{log}P_{\theta_1}(\mathbf{y}|\mathbf{x}) $\\
        $\theta_\mathrm{bottom\_k} \leftarrow \theta_\mathrm{bottom\_k} - \eta \bigtriangledown_{\theta_\mathrm{bottom\_k}}{l_1}  $
        
	}
}
\tcp{\color{mycolor} Stage 2, only $\theta_\mathrm{top\_m}$ being trainable.}
 \For {epoch $l = 1$ to $L$}   
{
	Shuffle $\mathcal{D}_{\mathrm{multi}}$ to obtain a new training sequence. \\
	\For {each batch $\mathcal{D}_2 \in \mathcal{D}_{\mathrm{multi}} $} 
	{ 
        $l_2 = \sum_{\mathbf{x,y}\sim\mathcal{D}_2} - \mathrm{log}P_{\theta_2}(\mathbf{y}|\mathbf{x})$  \\
        $\theta_\mathrm{top\_m} \leftarrow \theta_\mathrm{top\_m} - \eta \bigtriangledown_{\theta_\mathrm{top\_m}}{l_2} $
        
	}
}
\caption{\small \modelseries Two-Stage Training.}
\label{alg:training_recipe}
\end{algorithm}

%% file: ACL_2026/sections/4.experiments.tex
\input{ACL_2026/images/lora_fft_compare}

\section{Experiments}

\subsection{Setting}

\input{ACL_2026/tabs/translation_flores_multilingual}

\paragraph{Models and Baselines} 
We compare \modelseries with a range of baselines across four categories:
(1) General instruction models, including the Gemma series~\cite{teamGemma2Improving2024,teamGemma3Technical2025}, Llama3 series~\cite{touvron2023llama2openfoundation, dubeyLlama3Herd2024}, and Qwen series~\cite{qwenQwen25TechnicalReport2025,yangQwen3TechnicalReport2025a};
(2) Different tuning strategies on instruction models using our parallel data, including full fine-tuning and LoRA tuning.
(3) Multilingual-enhanced models, including the Tower series~\cite{alvesTowerOpenMultilingual2024}, Aya series~\cite{dangAyaExpanseCombining2024, ustun2024aya}, Sailor2~\cite{douSailor2SailingSouthEast2025}, and LLaMAX3-8B-Alpaca~\cite{lu-etal-2024-llamax};
(4) Domain-specialized LLMs, such as the Qwen2.5-Math~\cite{yangQwen25MathTechnicalReport2024}, and the Qwen2.5-Coder~\cite{huiQwen25CoderTechnicalReport2024}.



\paragraph{Evaluation datasets and Metrics} 
We evaluate \modelseries on FLORES-101~\cite{flores101} using the BenchMAX~\cite{benchmax:2025} evaluation suitcase~\footnote{\url{https://huggingface.co/datasets/LLaMAX/BenchMAX_General_Translation}}. For translation evaluation, we adopt two metrics: spBLEU~\cite{flores101} and xComet~\cite{guerreiro-etal-2024-xcomet}. The spBLEU metric measures translations based on text surface, while xComet focuses on the semantic similarity between the source sentence and the translation. By using both metrics, we avoid inflated xComet scores that can arise from directly copying the source sentence, while also accounting for different valid translation possibilities.

\input{ACL_2026/images/complex_problem}

\input{ACL_2026/tabs/multilingual_tasks}

For our multilingual tasks, we evaluate seven benchmarks—XNLI~\cite{conneau-etal-2018-xnli}, MGSM~\cite{shiLanguageModelsAre2022b,benchmax:2025}, xIFEval~\cite{benchmax:2025}, XStoryCloze~\cite{lin-etal-2022-shot}, XCOPA~\cite{ponti-etal-2020-xcopa}, XGPQA~\cite{benchmax:2025,rein2024gpqa}, and XWinograd~\cite{muennighoff-etal-2023-crosslingual}—using the BenchMax suite and the lm-evaluation-harness~\cite{eval-harness} suite to assess the accuracy metrics for each task.

For popular reasoning tasks, we evaluate 15 benchmarks, including BBEH~\cite{kazemi-etal-2025-big}, AIME2024~\cite{ArtProblemSolving}, AIME2025~\cite{ArtProblemSolving}, OlympiadBench~\cite{he-etal-2024-olympiadbench}, LiveMathBench~\cite{liu-etal-2025-llms-capable}, OlymMath~\cite{sunChallengingBoundariesReasoning2025}, Math~\cite{austinProgramSynthesisLarge2021},  LiveCodeBench-V5, LiveCodeBench-V6~\cite{jainLiveCodeBenchHolisticContamination2024a}, BigCodeBench-Hard~\cite{zhuoBigCodeBenchBenchmarkingCode2024},
and HumanEval~\cite{chenEvaluatingLargeLanguage2021}.
We utilize the OpenCompass~\cite{2023opencompass} suite, employing accuracy as the metric for all tasks except for LiveCodeBench-V5, LiveCodeBench-V6, BigCodeBench-Hard, 
and HumanEval, which use pass@1 as the metric.

\subsection{Experimental Results}
\subsubsection{Effectiveness of \method}

In Figure~\ref{fig:lora_fft_compare}, we compared the performance of \modelseries with its start model, Qwen3, as well as two alternative fine-tuning strategies: full fine-tuning (FFT) and LoRA.
Since Qwen3-8B and Qwen3-14B are instruction-tuned models, enhancing their capabilities with limited data is non-trivial, and FFT on these models often causes catastrophic forgetting. For example, on Qwen3-8B, FFT leads to degraded translation performance across most languages.
In comparison, LoRA helps mitigate catastrophic forgetting, but even after LoRA training, the model’s translation performance in most languages still falls short of its start model, Qwen3-8B or Qwen3-14B.
In contrast, \method effectively improves the translation performance of Qwen3. Across 28 experimental settings, \modelseries achieved higher xComet scores than Qwen3 in 27 cases. The improvement is especially pronounced for weaker languages like \texttt{sw}.
Complete results are shown in Table~\ref{tab:translation_flores_qwen3}.



\subsubsection{Performance Comparison}

We evaluate \modelseries against a range of advanced LLMs on translation, multilingual, and general tasks and finds that \modelseries exhibits the following strengths:

\paragraph{Superior Translation Capability} As shown in Table~\ref{tab:translation_flores_multilingual}, \modelseries demonstrates leading translation performance among current top-performing multilingual LLMs.
In both the many-to-one and one-to-many settings, \modelseries achieves the highest xComet scores in 6 of the 7 reported languages.
Notably, \modelseries outperforms even larger models. Despite using less than half the parameters, \modellarge achieves higher xComet scores than Aya-Expanse-32B across all evaluated languages, and \modelsmall, with only one quarter of the parameters, surpasses Aya-Expanse-32B in 12 of the 14 translation directions.
Moreover, the advantage of \modelseries is particularly pronounced for low-resource languages, where \modellarge outperforms the second-best model, LLaMAX3-8B-Alpaca, by 4.51, 8.04, and 10.67 xComet scores on $\text{x} \rightarrow \text{sw}$, $\text{x} \rightarrow \text{th}$, and $\text{x} \rightarrow \text{bn}$, respectively.
Complete results are shown in Table~\ref{tab:translation_flores}.


\paragraph{Improved Multilingual Capability} In Table~\ref{tab:multilingual_tasks}, we evaluate \modelseries on 7 multilingual datasets. Despite being trained solely on general parallel corpora without any task-specific multilingual data, \modelseries demonstrates improved multilingual capabilities compared to its start models. Specifically, \modelsmall outperforms Qwen3-8B on 6 out of 7 datasets, while \modellarge outperforms Qwen3-14B on 5 out of 7 datasets.
Furthermore, compared to other multilingual LLMs, \modelseries consistently ranks among the best across all evaluated datasets. Its performance is particularly strong on xIFEval and XGPQA, where it exceeds the scores of existing top-performing multilingual models.
Complete results are shown in Table~\ref{tab:general_task}.

\paragraph{Sustained General Reasoning Capability} Training instruction-tuned models on a single task often leads to forgetting of general capabilities. However, as shown in Figure~\ref{fig:general_tasks}, \modelseries maintains consistently stable general capabilities. Across reasoning tasks, including mathematics and code, \modelseries consistently performs on par with its start model. Notably, compared to the current leading multilingual model Tower-Plus-9B, \modelseries demonstrates a clear advantage.


\input{ACL_2026/tabs/layer_combination_analysis}

\subsubsection{Core Findings}

We investigate three key factors underlying the strong performance of \modelseries.

\paragraph{The Start Model}
\modelseries aligns into the multilingual space starting from an instruct model rather than a base model, thereby leveraging the stronger capabilities of the Instruct variant. Traditional multilingual models usually rely on base models for continued pre-training, assuming better transferability, but this overlooks the substantial abilities already embedded in Instruct models, which are trained on extensive high-quality instruction-tuning corpora, much of it non-public. 
For instance, Sailor2-8B-Chat, built on Qwen2.5-7B-Base, shows weaker translation performance than the Instruct version Qwen2.5-7B and even lags behind the domain-specialized LLM Qwen2.5-Coder-7B, as can be observed in Table~\ref{tab:translation_flores}. In contrast, by employing \method, \modelseries achieves a smooth training process from instruct models, inheriting their strengths while further extending multilingual capability.

\input{ACL_2026/tabs/one_stage_vs_two_stage}

\paragraph{The Data Requirements}
By starting from Instruct models, \modelseries demonstrates strong performance across diverse tasks and aligns multilingual capability using only a small amount of data, without relying on massive data for capability enhancement. Specifically, whereas Hunyuan-MT uses 1.3T tokens, Sailor2 uses 500B tokens, and Tower Plus uses 32B tokens, \modelseries attains the most competitive multilingual and general-task performance with only 0.8B tokens. In particular, although \modelseries is trained solely in general parallel corpora, it achieves highly competitive performance on specialized tasks such as code and math (Figure~\ref{fig:general_tasks}). However, for knowledge-intensive multilingual tasks like XGPQA (Table~\ref{tab:multilingual_tasks}), \modelseries does not surpass its start model, indicating that task-specific domain knowledge is essential for such tasks and cannot be fully compensated by general parallel corpora alone.

\paragraph{The Training Process} \modelseries is trained using an efficient training procedure. By relying solely on the SFT stage, it achieves the best multilingual capability among models of comparable scale, without requiring the more demanding CPT or RL phases. This demonstrates the effectiveness of our \method approach and indicates the potential for further improvements through the incorporation of denser training stages.





%% file: ACL_2026/images/lora_fft_compare.tex
\begin{figure*}[h]
    \centering
    \begin{subfigure}[b]{0.45\linewidth}
        \includegraphics[height=3.5cm]{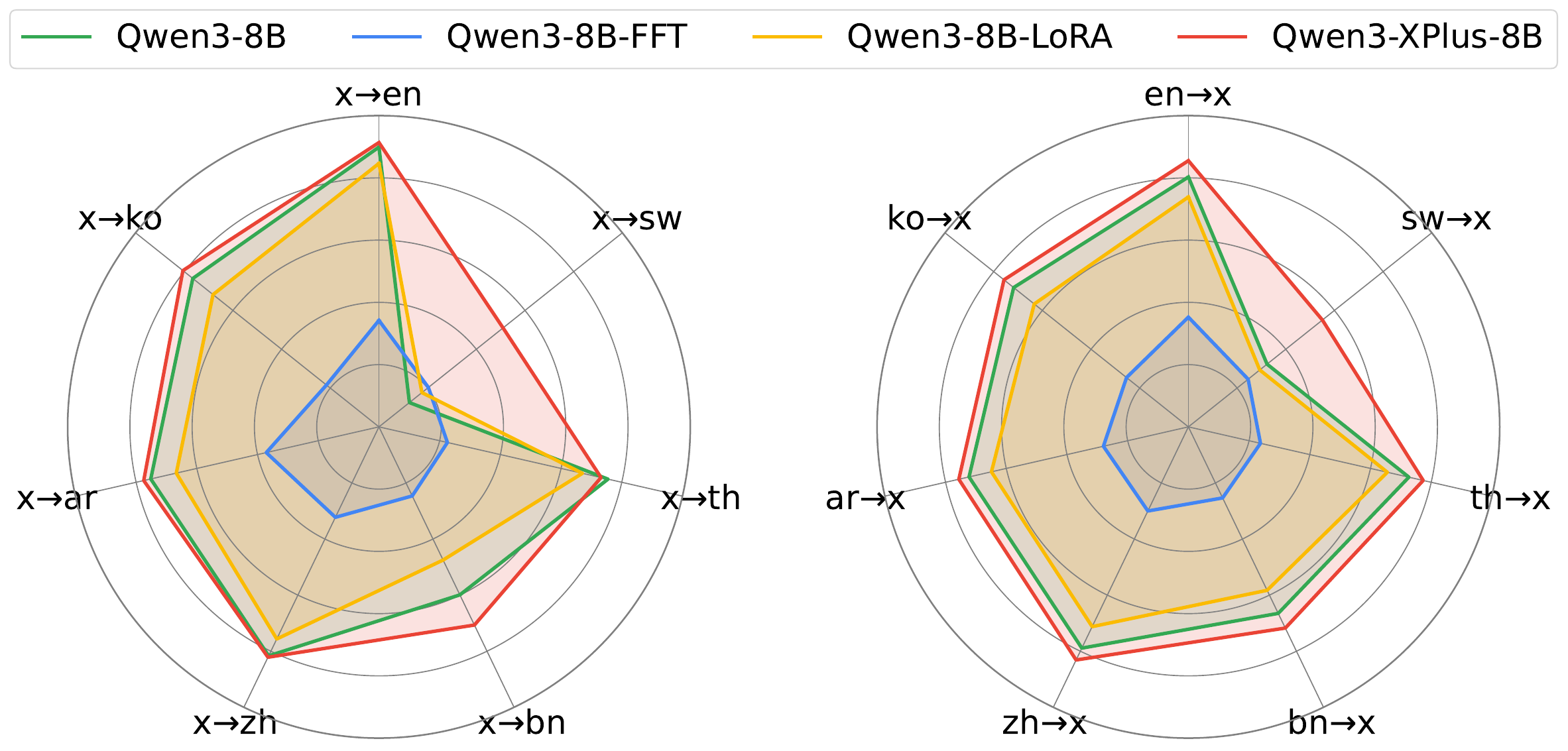}
        \caption{\modelsmall}
    \end{subfigure}
    \hspace{0.05\linewidth}
    \begin{subfigure}[b]{0.45\linewidth}
        \includegraphics[height=3.5cm]{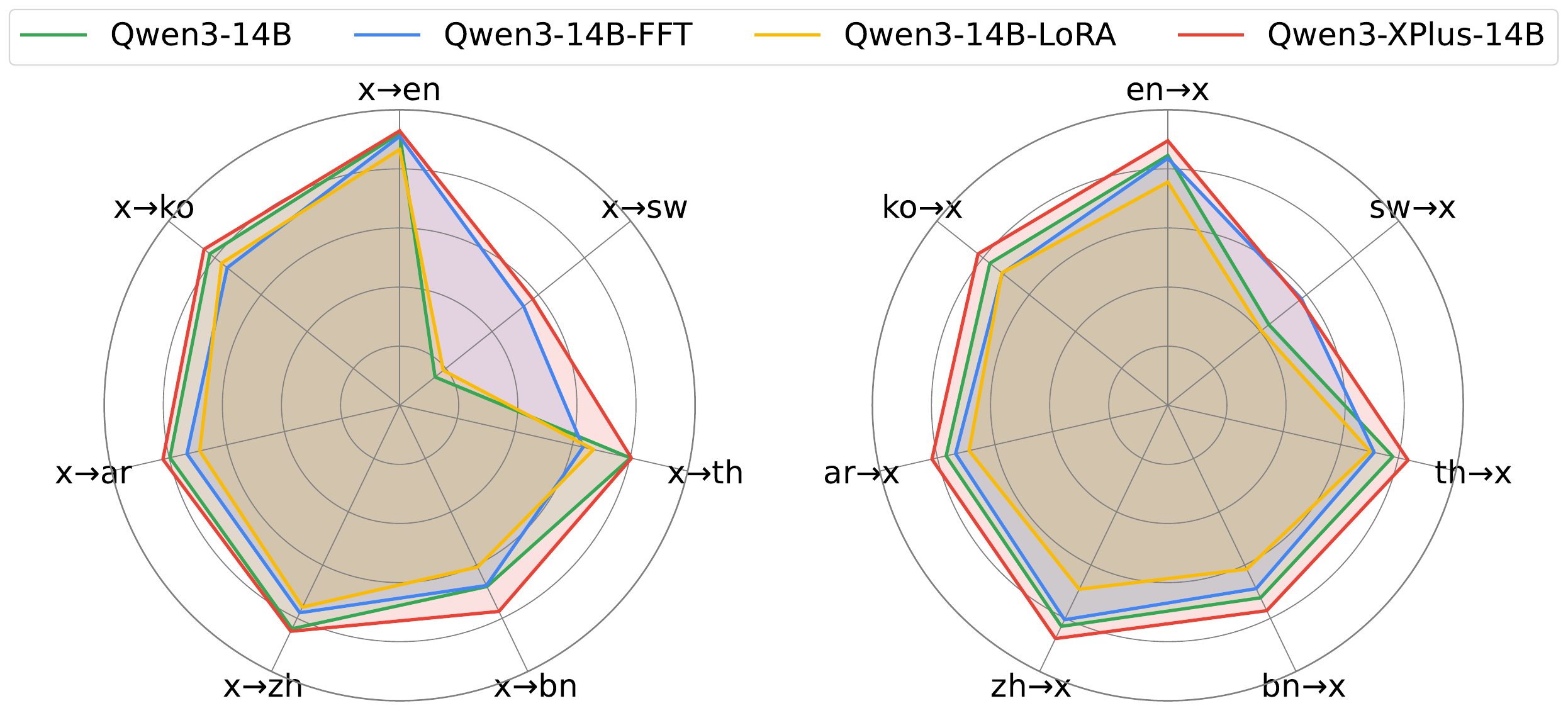}
        \caption{\modellarge}
    \end{subfigure}

    \caption{
        Comparison of xComet scores of \modelseries with Qwen3, and between Full Fine-Tuning~(FFT) and LoRA 
        on the FLORES-101 test set covering 17 languages, with results for 7 representative languages shown in the figure. 
        The results demonstrate that Layer-selective Tuning consistently enhances the translation performance of Qwen3 compared with both LoRA and FFT. In this figure, “x” denotes translation into any of the other 16 languages, excluding the source and target languages in each translation direction.
    }
    \label{fig:lora_fft_compare}
\end{figure*}

%% file: ACL_2026/tabs/translation_flores_multilingual.tex
\begingroup
\renewcommand{\arraystretch}{1} 
\begin{table*}[!ht]
\centering
\footnotesize
\resizebox{\linewidth}{!}{
\begin{tabular}{l|rr|rr|rr|rr|rr|rr|rr}

\toprule
& \multicolumn{2}{c|}{\textbf{x $\rightarrow$ en}} & \multicolumn{2}{c|}{\textbf{x $\rightarrow$ sw}} & \multicolumn{2}{c|}{\textbf{x $\rightarrow$ th}} & \multicolumn{2}{c|}{\textbf{x $\rightarrow$ bn}} & \multicolumn{2}{c|}{\textbf{x$\rightarrow$ zh}} & \multicolumn{2}{c|}{\textbf{x $\rightarrow$ ar}} & \multicolumn{2}{c}{\textbf{x $\rightarrow$ ko}} \\
& \textbf{spBLEU} & \textbf{xComet} & \textbf{spBLEU} & \textbf{xComet} & \textbf{spBLEU} & \textbf{xComet} & \textbf{spBLEU} & \textbf{xComet} & \textbf{spBLEU} & \textbf{xComet} & \textbf{spBLEU} & \textbf{xComet} & \textbf{spBLEU} & \textbf{xComet} \\

\midrule
 
TowerInstruct-7B-v0.1 & 29.26 & 72.56 & 0.71 & 37.34 & 0.48 & 53.90 & 0.25 & 59.70 & 17.02 & 62.16 & 0.48 & 58.62 & 13.4 & 62.28 \\
Hunyuan-MT-7B & 21.20 & 67.68 & 5.55 & 32.95 & 17.70 & 56.92 & 8.92 & 47.17 & 18.35 & 73.67 & 13.70 & 54.37 & 10.32 & 58.28 \\
Sailor2-8B-Chat & 0.52 & 17.81 & 0.67 & 17.09 & 24.12 & 60.84 & 2.42 & 20.67 & 16.69 & 60.53 & 5.60 & 31.41 & 4.46 & 37.03 \\
LLaMAX3-8B-Alpaca & 35.96 & 89.98 & 10.00 & 53.15 & 23.62 & 72.43 & 12.04 & 66.76 & 21.08 & 77.72 & 17.57 & 72.17 & 11.90 & 76.11 \\
Tower-Plus-9B & \textbf{40.12} & 91.74 & 2.45 & 20.80 & 18.71 & 53.76 & 2.47 & 58.16 & \textbf{30.37} & 82.96 & 9.66 & 48.73 & \textbf{22.36} & \textbf{85.53} \\
Aya-Expanse-8B & 33.13 & 79.28 & 1.49 & 8.91 & 6.42 & 19.81 & 4.94 & 25.08 & 23.53 & 70.67 & 23.77 & 70.21 & 17.71 & 70.53 \\
Aya-Expanse-32B & 39.72 & 88.63 & 2.60 & 16.53 & 15.16 & 40.65 & 11.93 & 53.76 & 27.93 & 80.70 & \textbf{28.63} & 81.70 & 21.71 & 82.79 \\

\midrule
Qwen3-8B & 35.24 & 89.89 & 3.49 & 12.52 & 29.85 & 75.47 & 14.14 & 59.91 & 26.88 & 81.65 & 21.73 & 75.18 & 16.20 & 76.48 \\
\modelsmall & 38.02 & 91.35 & 18.60 & 50.99 & 27.84 & 73.17 & 19.39 & 70.67 & 26.95 & 82.15 & 24.00 & 77.50 & 18.08 & 80.54 \\

\midrule
Qwen3-14B & 36.92 & 91.98 & 5.87 & 15.33 & 32.40 & 80.12 & 17.50 & 68.09 & 28.71 & 83.95 & 24.01 & 79.75 & 18.77 & 82.19 \\
\modellarge & 39.01 & \textbf{92.86} & \textbf{20.02} & \textbf{57.66} & \textbf{32.03} & \textbf{80.47} & \textbf{21.63} & \textbf{77.43} & 28.96 & \textbf{84.90} & 26.31 & \textbf{82.19} & 20.31 & 84.68 \\
\bottomrule

\toprule
& \multicolumn{2}{c|}{\textbf{en $\rightarrow$ x}} & \multicolumn{2}{c|}{\textbf{sw $\rightarrow$ x}} & \multicolumn{2}{c|}{\textbf{th $\rightarrow$ x}} & \multicolumn{2}{c|}{\textbf{bn $\rightarrow$ x}} & \multicolumn{2}{c|}{\textbf{zh $\rightarrow$ x}} & \multicolumn{2}{c|}{\textbf{ar $\rightarrow$ x}} & \multicolumn{2}{c}{\textbf{ko $\rightarrow$ x}} \\
& \textbf{spBLEU} & \textbf{xComet} & \textbf{spBLEU} & \textbf{xComet} & \textbf{spBLEU} & \textbf{xComet} & \textbf{spBLEU} & \textbf{xComet} & \textbf{spBLEU} & \textbf{xComet} & \textbf{spBLEU} & \textbf{xComet} & \textbf{spBLEU} & \textbf{xComet} \\

\midrule

TowerInstruct-7B-v0.1 & 18.26 & 67.80 & 2.67 & 17.33 & 3.82 & 29.22 & 1.78 & 21.56 & 10.89 & 69.26 & 7.95 & 39.09 & 11.26 & 70.69 \\
Hunyuan-MT-7B & 28.43 & 87.96 & 14.12 & 39.86 & 7.53 & 36.37 & 4.60 & 28.71 & 20.37 & 83.94 & 15.72 & 55.10 & 14.31 & 51.93 \\
Sailor2-8B-Chat & 16.03 & 54.11 & 1.76 & 14.08 & 4.30 & 33.86 & 5.72 & 30.13 & 3.83 & 28.77 & 7.29 & 34.26 & 7.18 & 36.11 \\
LLaMAX3-8B-Alpaca & 26.62 & 83.22 & 20.23 & \textbf{59.06} & 16.03 & 74.88 & 17.20 & 68.49 & 16.51 & 80.33 & 18.37 & 73.77 & 17.39 & 72.23 \\
Tower-Plus-9B & 28.83 & 79.85 & 18.38 & 46.19 & 19.02 & 70.29 & 18.64 & 62.92 & 19.39 & 75.55 & 21.72 & 69.28 & 20.68 & 71.93 \\
Aya-Expanse-8B & 25.75 & 68.36 & 7.90 & 16.43 & 11.39 & 40.78 & 11.29 & 36.77 & 17.85 & 65.45 & 20.21 & 60.86 & 18.41 & 60.74 \\
Aya-Expanse-32B & 30.21 & 78.30 & 16.72 & 38.82 & 18.25 & 64.11 & 19.37 & 62.04 & 21.26 & 75.21 & 24.71 & 71.16 & 22.07 & 70.84 \\

\midrule

Qwen3-8B & 28.86 & 80.27 & 13.61 & 32.32 & 19.68 & 72.64 & 19.38 & 66.51 & 20.82 & 78.90 & 22.85 & 72.27 & 20.23 & 71.79 \\
\modelsmall & 32.82 & 85.52 & 21.41 & 55.06 & 22.68 & 77.36 & 22.47 & 71.75 & 23.31 & 83.13 & 25.46 & 75.63 & 22.73 & 75.77 \\
\midrule
Qwen3-14B & 31.78 & 84.54 & 18.40 & 43.75 & 22.35 & 78.09 & 22.47 & 72.32 & 23.02 & 83.04 & 25.28 & 76.98 & 22.90 & 77.07 \\
\modellarge & \textbf{35.97} & \textbf{89.51} & \textbf{23.19} & 57.23 & \textbf{24.91} & \textbf{83.40} & \textbf{24.93} & \textbf{77.15} & \textbf{25.41} & \textbf{87.64} & \textbf{28.18} & \textbf{81.90} & \textbf{25.53} & \textbf{82.16}  \\

\bottomrule

\end{tabular}%
}

\caption{Comparison translation performance of \modelseries with Qwen3, and between Full Fine-Tuning~(FFT) and LoRA on the FLORES-101 test set covering 17 languages, with results for 7 representative languages shown in the table.
\modellarge delivers the best performance on 21 of the 28 reported metrics. In this table, “x” denotes translation into any of the other 16 languages, excluding the source and target languages in each translation direction. In this table, “x” denotes translation into any of the other 16 languages, excluding the source and target languages in each translation direction.}
\label{tab:translation_flores_multilingual}
\end{table*}
\endgroup

%% file: ACL_2026/images/complex_problem.tex
\begin{figure*}[!h]
    \centering
    \includegraphics[width=\textwidth]{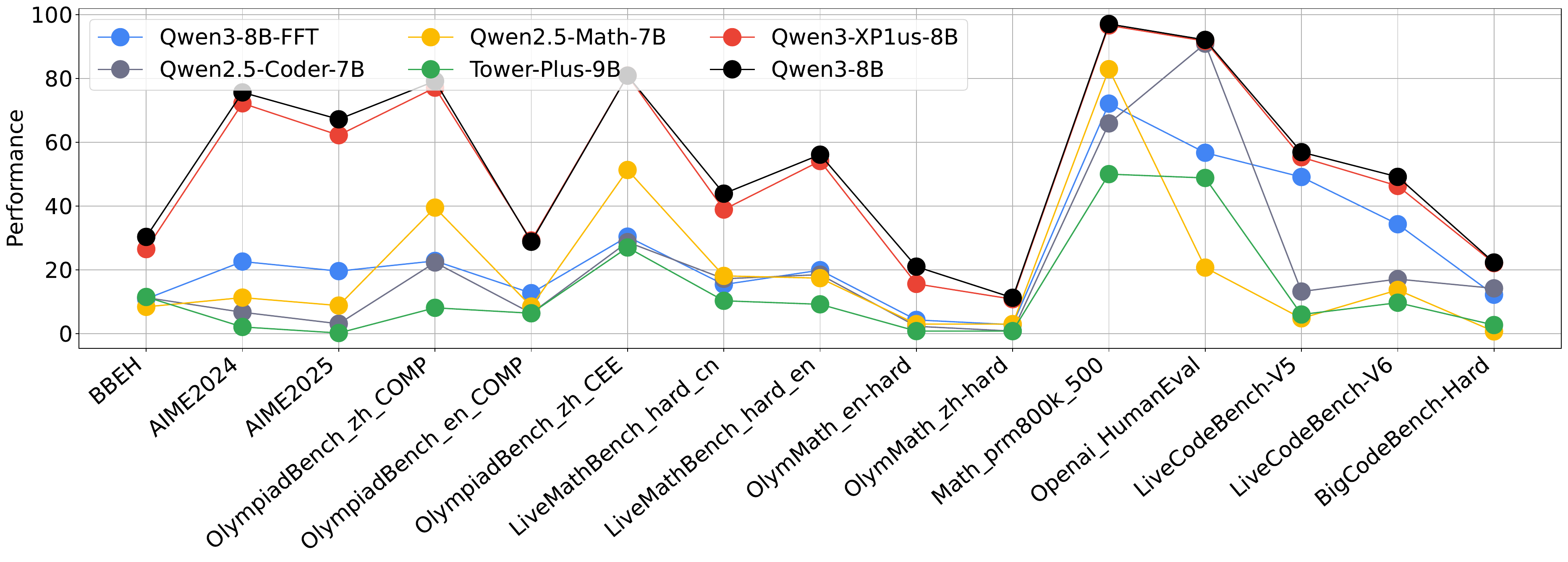}
    \caption{Comparison of \modelseries, its start model Qwen3,Qwen2.5-Math-7B, Qwen2.5-Coder-7B and the leading multilingual model Tower-Plus-9B on 15 reasoning datasets. The results show that \modelseries achieves overall performance comparable to Qwen3 and significantly surpasses Tower-Plus-9B.}
    \label{fig:general_tasks}
\end{figure*}

%% file: ACL_2026/tabs/multilingual_tasks.tex
\begin{table*}[!t]
\centering
\footnotesize

\begin{tabular}{@{}l|c|c|c|c|c|c|c|c@{}}
\toprule
\textbf{Models} &\textbf{ AVG.} & \textbf{XNLI} & \textbf{MGSM} & \textbf{xIFEval} & \textbf{XStoryCloze} & \textbf{XCOPA} & \textbf{XGPQA} & \textbf{XWinograd} \\
\midrule



Qwen3-8B & 55.30 & 42.44 & 44.87 & \textbf{81.68} & 58.08 & 60.40 & \textbf{35.85} & 63.79 \\
\modelsmall & \textbf{56.93} & \textbf{44.79} & \textbf{50.36} & 80.49 & \textbf{59.24} & \textbf{61.44} & 34.26 & \textbf{67.95} \\
\midrule
Qwen3-14B & 57.64 & 43.05 & \textbf{52.18} & \textbf{85.64} & 58.73 & 61.87 & \textbf{41.43} & 60.58 \\
\modellarge & \textbf{58.61} & \textbf{44.77} & 50.22 & 85.55 & \textbf{61.14} & \textbf{63.60} & 40.10 & \textbf{64.87} \\



\bottomrule
\end{tabular}%
\caption{Comparison of \modelseries and Qwen3 on 7 multilingual tasks. Using only general parallel corpora and no task-specific multilingual data, \modelseries  wins 5 out of 7 tasks against Qwen3.}
\label{tab:multilingual_tasks}
\end{table*}

%% file: ACL_2026/tabs/layer_combination_analysis.tex
\begin{table}[!t]
\centering
\small
\begin{tabular}{@{}c|c|cc@{}}
\toprule
\textbf{Setting} & \textbf{Layer} & \textbf{en$\rightarrow$x} & \textbf{x$\rightarrow$en} \\
\midrule
\multicolumn{2}{c|}{Qwen3-8B} & 28.91 & 35.26 \\
\midrule
\multirow{3}{*}{Bottom} & 4 & 36.53 & 29.64 \\
 & 8 & 33.55 & 22.00 \\
 & 15 & 27.93 & 17.79 \\
 \midrule
\multirow{3}{*}{Top} & 4 & 25.98 & 33.79 \\
 & 8 & 27.89 & 35.25 \\
 & 15 & 28.92 & 36.39 \\
 \midrule
\multirow{3}{*}{Bottom 4 + Top} & 4 & 32.28 & 36.40 \\
 & 8 & 33.12 & 36.59 \\
 & 15 & 32.82 & 38.02 \\
 \bottomrule
\end{tabular}%
\caption{Ablation study on layer selection.}
\label{tab:layer_ablation}
\end{table}

%% file: ACL_2026/tabs/one_stage_vs_two_stage.tex
\begin{table}[]
\centering
\footnotesize
\resizebox{\linewidth}{!}{%
\begin{tabular}{@{}l|cc|cc|cc@{}}
\toprule
\multirow{2}{*}{\textbf{x}}  & \multicolumn{2}{c|}{\textbf{Qwen3-8B}} & \multicolumn{2}{c|}{\textbf{One-Stage}} & \multicolumn{2}{c}{\textbf{Two-Stage}} \\
& \textbf{x$\rightarrow$en} & \textbf{en$\rightarrow$x} & \textbf{x$\rightarrow$en} & \textbf{en$\rightarrow$x} & \textbf{x$\rightarrow$en} & \textbf{en$\rightarrow$x} \\
\midrule
ar & 38.28 & 29.38 & 42.09 & 31.98 & \textbf{42.17} & \textbf{33.03} \\
bn & 32.15 & 19.02 & \textbf{34.63} & 24.85 &  34.43 & \textbf{25.42}  \\
cs & 40.48 & 31.36 & 42.20 & 32.44 &  \textbf{42.88} & \textbf{32.99}  \\
de & 44.73 & 40.44 & \textbf{46.76} & \textbf{40.21} & 46.65 & 39.99 \\
es & 32.63 & 30.21 & \textbf{34.14} & 31.12 &  33.98 & \textbf{31.43}  \\
fr & 46.00 & 49.44 & 47.52 & 50.32 & \textbf{47.60} & \textbf{50.73} \\
hu & 35.59 & 24.09 & 37.82 & 25.41 &  \textbf{38.03} & \textbf{27.39}  \\
ja & 29.04 & 24.79 & 31.22 & 25.45 &  \textbf{31.33} & \textbf{26.11}  \\
ko & 31.44 & 21.13 & \textbf{32.94} & 22.48 &  32.72 & \textbf{23.30}  \\
ru & 36.56 & 35.19 & \textbf{38.39} & 35.11 &  38.18 & \textbf{36.23}  \\
sr & 41.10 & 25.23 & 43.57 & 32.00 &  \textbf{44.19} & \textbf{32.50}  \\
{\color[HTML]{D83931} sw} & {\color[HTML]{D83931} 23.08} & {\color[HTML]{D83931} 6.53} & {\color[HTML]{D83931} \textbf{32.03}} & {\color[HTML]{D83931} 23.32} & {\color[HTML]{D83931} \textbf{33.55} } & {\color[HTML]{D83931} \textbf{26.53}} \\
{\color[HTML]{D83931} te} & {\color[HTML]{D83931} 33.51} & {\color[HTML]{D83931} 15.83} & {\color[HTML]{D83931} 36.18} & {\color[HTML]{D83931} 27.54} & {\color[HTML]{D83931} \textbf{36.63} } & {\color[HTML]{D83931} \textbf{30.13}} \\
th & 31.17 & 36.53 & \textbf{33.04} & 32.91 &  32.99 & \textbf{33.76}  \\
vi & 37.34 & 38.34 & \textbf{39.80} & 40.30 &  39.36 & \textbf{40.80}  \\
zh & 31.10 & 34.99 & 33.35 & \textbf{35.68} &   \textbf{33.65} & 34.71  \\
\midrule
Avg & 35.26 & 28.91 & 37.86 & 31.95 & \textbf{38.02} & \textbf{32.81} \\
\bottomrule
\end{tabular}%
}
\caption{Effectiveness of two-stage tuning in \method. Experimental results show that the two-stage tuning strategy offers advantages over the one-stage approach and brings pronounced benefits for low-resource languages such as sw and te.}
\label{tab:two_vs_one}
\end{table}

%% file: ACL_2026/sections/5.analysis.tex
\section{Analysis}

\subsection{Layer Combination Analysis}

\input{ACL_2026/images/figure_unseen_languages}

Furthermore, Table~\ref{tab:layer_ablation} investigates the impact of different layer selection strategies in \method on model performance. We first evaluate training exclusively on the lower layers and exclusively on the higher layers, and then examine combinations of both.

Training a few of the lower layers already surpasses the baseline, with the bottom four layers achieving the best results. For the higher layers, training the top fifteen achieves the largest improvement, likely because they capture more complex semantic features. Notably, layer 20 (the 16th from the top) negatively impacts performance and is skipped in the current experiments (Figure~\ref{fig:single_lyr_tuning}). 

Finally, when combining lower and higher layers, we observe that training the bottom four together with the top fifteen layers delivers the best translation performance. Consequently, this configuration is adopted in our main experiments.

\subsection{Effect of Two-Stage Training}

In \method, the training process is designed in two stages to effectively adapt different layers of the model. To evaluate the necessity of this two-stage design, we compared it with a single-stage approach in Table~\ref{tab:two_vs_one}. 
The results show that sequentially fine-tuning the lower layers followed by the higher layers provides advantages over training both simultaneously. This improvement is likely due to the smoother adaptation process afforded by the two-stage design.
Notably, even single-stage training significantly outperforms the start model Qwen3-8B, highlighting the importance of carefully selecting layers for fine-tuning in \method.

\input{ACL_2026/images/llama3.1_next}

\subsection{Generalization to Unseen Languages}

\modelseries is trained on parallel corpora covering 17 languages. To evaluate its generalization to unseen languages, we test on 12 representative ones (Figure~\ref{fig:unseen_languages}). The results show that \modellarge consistently outperforms Qwen3-8B, demonstrating robust cross-lingual generalization and confirming the method’s effectiveness in extending multilingual ability beyond the training set.

\subsection{Adaptability to Different Backbones}

To verify the generality of \method across different models, we apply it to Llama3.1-8B using the same training setup. As shown in Figure~\ref{fig:llama3.1_next}, \method substantially improves performance across multiple languages, with particularly notable gains on low-resource languages. These results demonstrate the broad applicability and potential of our approach.

\subsection{Adaptability to Different Tasks}

In our main experiments, we apply \method on a translation training set.
To further evaluate its applicability beyond translation, we conduct experiments on code generation tasks and present the results in Table~\ref{tab:code_sft}.
Specifically, we fine-tune Qwen3-8B on two datasets: (1) python-related samples selected from OpenThoughts~\cite{OpenThoughts}, and (2) web-oriented samples synthesized to construct the WebSyn dataset.
We then compare the performance of Full Fine-Tuning (FFT) with our proposed \method. 

Experimental results show that \method consistently outperforms FFT on both OpenThoughts and WebSyn. On OpenThoughts, \method achieves 1.22\%--2.86\% higher accuracy, and on WebSyn it yields 0.61\%--4.00\% improvement. Notably, Qwen3-8B fine-tuned with \method gains 0.68\%--2.29\% across four WebSyn evaluation sets, while FFT decreases performance on three. These results indicate that \method remains effective even for code generation tasks.

\input{ACL_2026/tabs/code_pft}

%% file: ACL_2026/images/figure_unseen_languages.tex
\begin{figure}[!t]
    \centering
    \includegraphics[width=\linewidth]{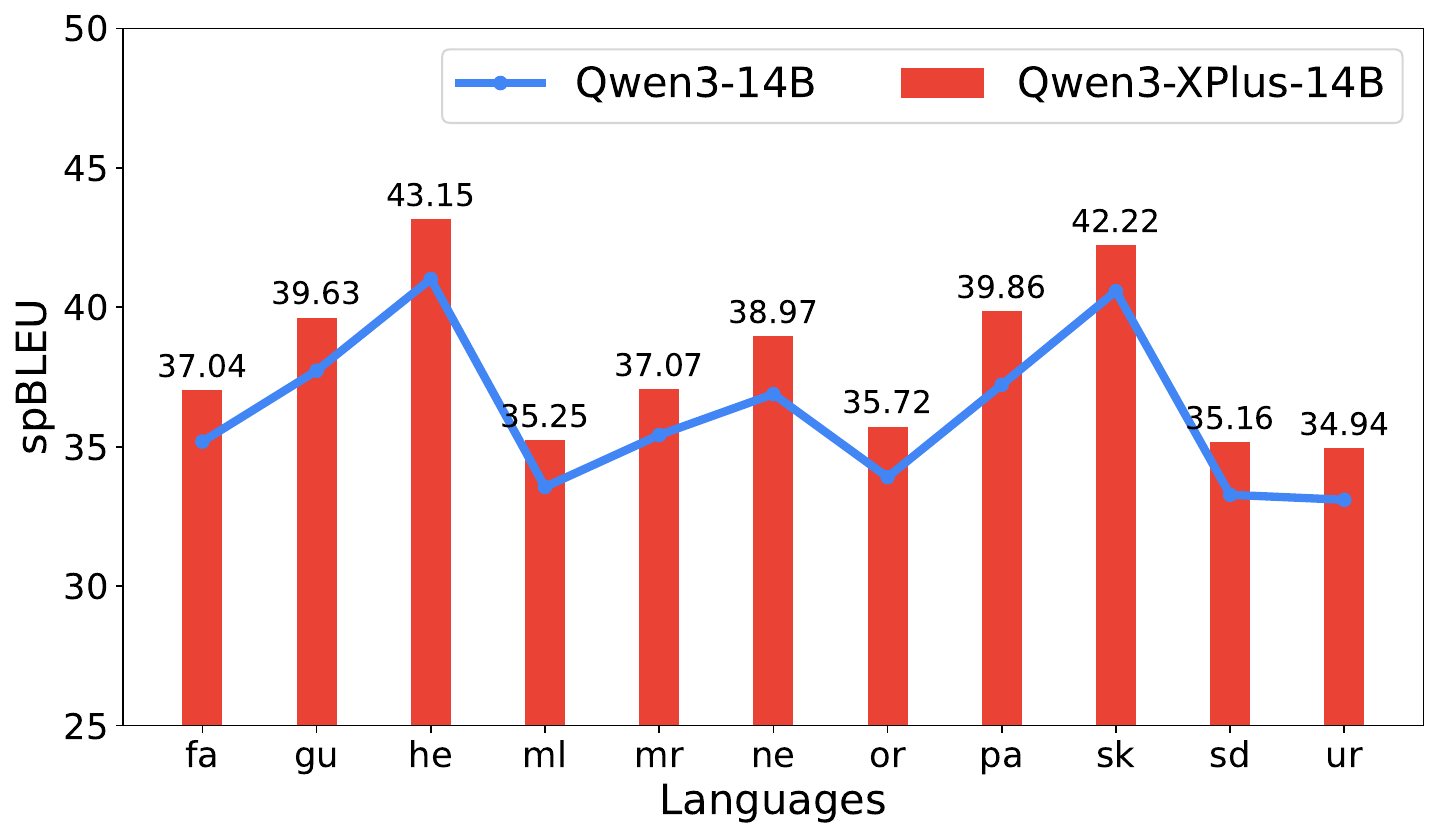}
    \caption{Translation performance on unseen languages. \modelseries also delivers gains on languages that were unseen during the \method stage.}
    \label{fig:unseen_languages}
\end{figure}

%% file: ACL_2026/images/llama3.1_next.tex
\begin{figure}[!t]
    \centering
    \includegraphics[width=\linewidth]{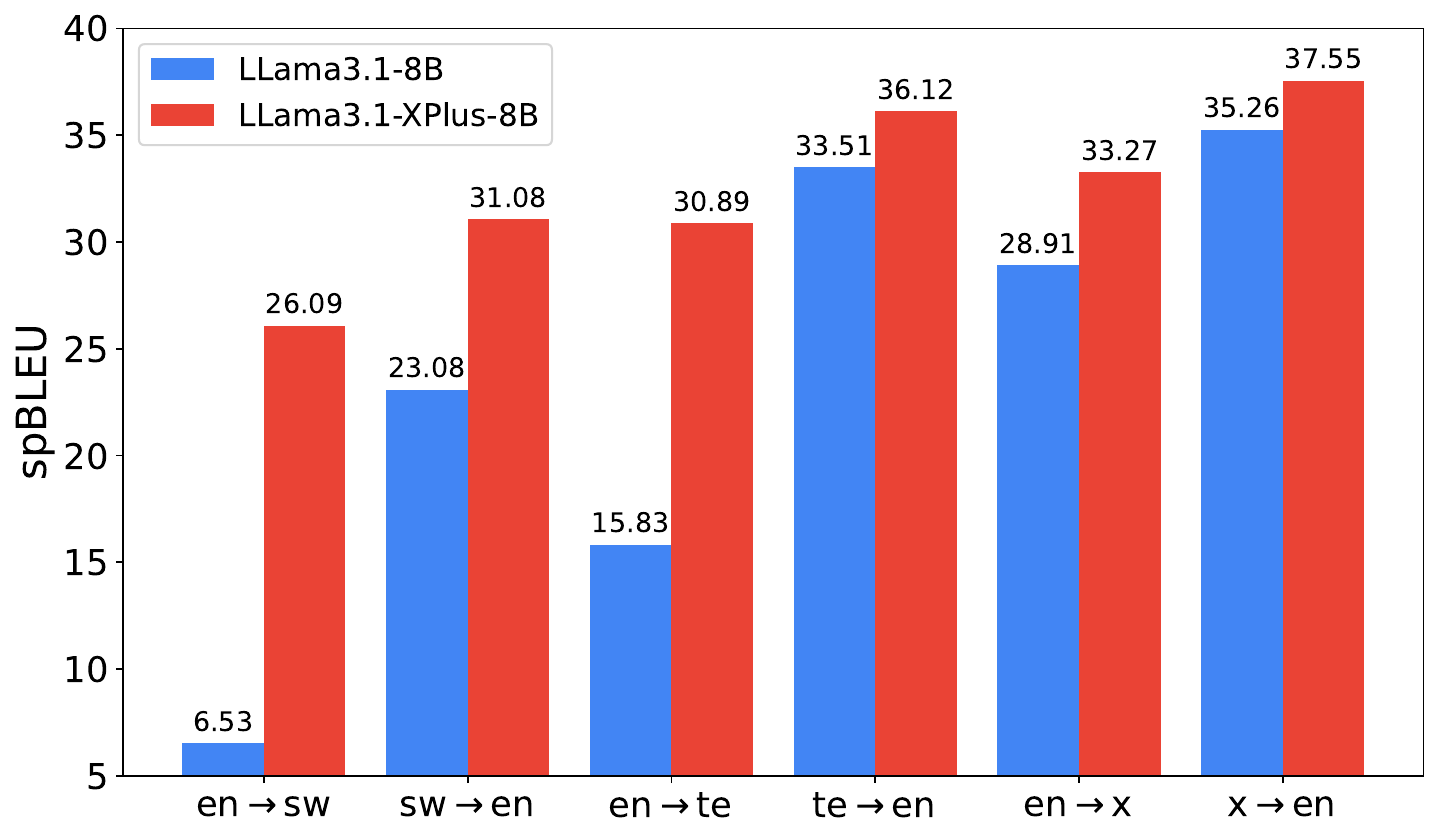}
    \caption{Experiments on different backbones. \method also brings improvements on the Llama-3.1-8B instruction model.}
    \label{fig:llama3.1_next}
\end{figure}

%% file: ACL_2026/tabs/code_pft.tex
\begin{table}[]
\centering
\resizebox{\linewidth}{!}{%
\begin{tabular}{@{}l|c|cc|cc@{}}
\toprule
\multirow{2}{*}{\textbf{Benchmark}} & \multirow{2}{*}{\textbf{Qwen3-8B}} & \multicolumn{2}{c|}{\textbf{OpenThoughts}} & \multicolumn{2}{c}{\textbf{WebSyn}} \\
& & \textbf{FFT} & \textbf{LT} & \textbf{FFT} & \textbf{LT} \\ 
\midrule
HumanEval & 92.68 & 81.71 & 82.93 & 93.29 & \textbf{93.90} \\
LiveCodeBench-V5 & 55.69 & 23.95 & 25.15 & 53.29 & \textbf{56.89} \\
LiveCodeBench-V6 & 48.57 & 23.43 & 26.29 & 46.86 & \textbf{50.86} \\
BigCodeBench-Hard & 25.00 & 18.92 & 20.27 & 22.30 & \textbf{25.68} \\
\bottomrule
\end{tabular}%
}
\caption{Comparison of Full Fine-Tuning (FFT) and our \method (LT) on Code Generation Tasks. In most cases, FFT leads to performance degradation in instruction models, whereas LT enhances their code generation capability.}
\label{tab:code_sft}
\end{table}

%% file: ACL_2026/sections/6.conclusion.tex
\section{Conclusion}

Our approach using \modelseries models demonstrates a significant enhancement in translation performance across diverse languages, particularly in low-resource contexts. By leveraging instruct models trained on limited parallel datasets, we achieve substantial gains in multilingual tasks while maintaining competitive reasoning capabilities. This research not only addresses the challenges faced by translation-enhanced models but also sets the stage for future developments in multilingual.

%% file: ACL_2026/sections/7.appendix.tex
\section{Appendix}
\input{ACL_2026/tabs/translation_flores_qwen3}
\input{ACL_2026/tabs/translation_flores}
\input{ACL_2026/tabs/understanding_results}

\subsection{Models}
Information of the models evaluated in our study are listed in Table \ref{tab:appendix-models}.

\begin{table*}[ht]
\centering
\scriptsize
\begin{tabular}{@{}c|ccc@{}}
\toprule
Group & Model Name & Parameter Size & Introduction \\ \midrule
\multirow{5}{*}{\begin{tabular}[c]{@{}c@{}}General\\ Instruct\end{tabular}} & Gemma3-IT \cite{teamGemma3Technical2025} & 12B & SOTA multimodal open model from Google \\
 & LLaMA3-Instruct \cite{dubeyLlama3Herd2024} & 8B & Popular, classic open LLM from Meta \\
 & LLaMA3.1-Instruct \cite{dubeyLlama3Herd2024} & 8B & Updated version of LLaMA3-Instruct \\
 & Qwen2.5-Instruct \cite{qwenQwen25TechnicalReport2025} & 8B, 14B, 32B & Popular, classic open LLM from Alibaba \\
 & Qwen3 \cite{yangQwen3TechnicalReport2025a} & 8B, 14B & SOTA open LLM with mixed thinking mode from Alibaba \\ \midrule
\multirow{5}{*}{\begin{tabular}[c]{@{}c@{}}Domain-\\ Specialized\end{tabular}} & CodeLLaMA \cite{roziereCodeLlamaOpen2024} & 7B & Open code LLM based on LLaMA2 \\
 & InternLM2-Math \cite{caiInternLM2TechnicalReport2024a} & 7B & Open math LLM based on InternLM2 \\
 & DeepSeek-Coder-V2-Lite \cite{deepseek-aiDeepSeekCoderV2BreakingBarrier2024} & 16B & Open code LLM based on DeepSeek-V2 \\
 & Qwen2.5-Math \cite{yangQwen25MathTechnicalReport2024} & 7B & Open math LLM based on Qwen-2.5 \\
 & Qwen2.5-Coder \cite{huiQwen25CoderTechnicalReport2024} & 7B, 14B, 32B & Open code LLM based on Qwen-2.5 \\ \midrule
\multirow{6}{*}{\begin{tabular}[c]{@{}c@{}}Multilingual-\\ Enhanced\end{tabular}} & TowerInstruct-v0.1 \cite{alvesTowerOpenMultilingual2024a} & 7B & Multilingual translation LLM based on LLaMA2 \\
 & Hunyuan-MT \cite{zhengHunyuanMTTechnicalReport2025} & 7B & Multilingual translation model based on Hunyuan \\
 & Aya-Expanse \cite{dangAyaExpanseCombining2024} & 8B & Advanced multilingual LLM based on Command R \\
 & Sailor2-Chat \cite{douSailor2SailingSouthEast2025} & 8B & South-East Asia languages focused LLM based on Qwen2 \\
 & LLaMAX3-Alpaca \cite{lu-etal-2024-llamax} & 8B & Multilingual translaation LLM based on LLaMA3 \\
 & Tower-Plus \cite{reiTowerBridgingGenerality2025} & 9B & Multilingual translation and general LLM based on Gemma2 \\ \midrule
\multirow{4}{*}{\begin{tabular}[c]{@{}c@{}}Super-\\ Large\end{tabular}} & Qwen3 \cite{yangQwen3TechnicalReport2025a} & 235B (22B Active) & SOTA open LLM with mixed thinking mode from Alibaba \\
 & DeepSeek-V3 \cite{deepseek-aiDeepSeekV3TechnicalReport2025} & 671B (37B Active) & Popular open LLM from DeepSeek \\
 & DeepSeek-R1 \cite{deepseek-aiDeepSeekR1IncentivizingReasoning2025a} & 671B (37B Active) & Popular, classic open reasoning LLM from DeepSeek \\
 & Kimi-K2 \cite{teamKimiK2Open2025} & 1T (32B Active) & SOTA reasoning LLM from Moonshot \\ \midrule
\multirow{3}{*}{Ours} & Qwen3-FFT & 8B, 14B & Qwen3 with fully finetuning on our multilingual data \\
 & Qwen3-LoRA & 8B, 14B & Qwen3 with LoRA finetuning on our multilingual data \\
 & Qwen3-XPlus & 8B, 14B & Qwen3 with \method on our multilingual data \\ \bottomrule
\end{tabular}
\caption{Information of models used in our study.}
\label{tab:appendix-models}
\end{table*}

\subsection{Training Data}
Our training data mainly sources from NLLB and OPUS-100. Here we briefly introduce these two datasets.

\paragraph{NLLB.} Provided in CCMatrix \cite{schwenk-etal-2021-ccmatrix}, this dataset was created based on metadata for mined parallel corpus released by Meta AI \cite{nllbteam2022languageleftbehindscaling}. It contains parallel text for 148 English-centric and 1465 non-English-centric language pairs, with a complete size of ~450GB.

\paragraph{OPUS-100.} OPUS-100 is an English-centric multilingual corpus covering 100 languages. The languages were selected based on the volume of parallel data available in OPUS (\url{https://opus.nlpl.eu}). OPUS-100 contains approximately 55M sentence pairs. Of the 99 language pairs, 44 have 1M sentence pairs of training data, 73 have at least 100k, and 95 have at least 10k.

\subsection{Evaluation Benchmarks}
Table \ref{tab:appendix-benchmarks} lists the benchmarks used in our evaluation.

\subsection{Hyperparameter Settings}
We train the model for 1 epoch with a learning rate of 1e-5, scheduled by a cosine scheduler with a minimum learning rate of 2e-6 and a warmup ratio of 0.03. Mixed precision training (bf16) is used to improve efficiency. All experiments are performed on 8 NVIDIA H800 GPUs with a per-device training batch size of 1 and gradient accumulation over 2 steps\footnote{More training details can be found in the configuration file: \url{https://huggingface.co/LLaMAX/Qwen3-XPlus-17langs-14B/blob/main/training.yaml}}.

\subsection{Evaluation Benchmarks}

\begin{table*}[ht]
\centering
\scriptsize
\begin{tabular}{@{}c|ccl@{}}
\toprule
\textbf{Group} & \textbf{Benchmark Name} & \textbf{Metric} & \textbf{Information} \\ \midrule
\multirow{1}{*}{Translation} & FLORES-101 \cite{goyal-etal-2022-flores} & \multirow{1}{*}{spBLEU, xCOMET} & Parallel sentences for 101 languages extracted from English Wikipedia \\
 \midrule
\multirow{7}{*}{Multilingual} & XNLI \cite{conneau-etal-2018-xnli} & \multirow{7}{*}{Accuracy} & Subset of MNLI translated into 14 languages, about textual entailment \\
 & MGSM \cite{shiLanguageModelsAre2022b} &  & Subset of GSM translated into 10 languages, about grad-school math \\
 & xIFEval \cite{benchmax:2025} &  & IFEval translated into 17 languages, about instruction following \\
 & XStoryCloze \cite{lin-etal-2022-shot} &  & English StoryCloze translated into 10 languages, about story continuation \\
 & XCOPA \cite{ponti-etal-2020-xcopa} &  & COPA translated into 11 languages, about commonsense reasoning \\
 & XGPQA \cite{benchmax:2025,rein2024gpqa} & & translated into 17 languages, about challenging scientific questions \\
 & XWinograd \cite{muennighoff-etal-2023-crosslingual} &  & Winograd enriched to 6 languages, about coreference resolution \\ \midrule
\multirow{14}{*}{General} & MathQA \cite{amini-etal-2019-mathqa} & \multirow{10}{*}{Accuracy} & Math word problems adapted from AQuA-RAT \\
 & BBEH \cite{kazemi-etal-2025-big} &  & Extra hard version of Big-Bench with newer and harder tasks \\
 & AIME 2024, 2025 \cite{ArtProblemSolving} &  & Problems from the American Invittional Mathematics Examination \\
 & OlympiadBench \cite{he-etal-2024-olympiadbench} &  & Olympiad-level bilingual multimodal math and physics promblems \\
 & LiveMathBench \cite{liu-etal-2025-llms-capable}&  & Challenging latest questions from mathematical competitions \\
 & OlymMath \cite{sunChallengingBoundariesReasoning2025} &  & Olympiad-level math problems in parallel English and Chinese \\
 & Math \cite{lightmanLetsVerifyStep2023} &  & Challenging competition math problems with full step-by-step solutions \\
 & MBPP \cite{austinProgramSynthesisLarge2021} & \multirow{3}{*}{Pass@1} & Crowd-sourced entry level Python programming problems \\
 & LiveCodeBench-V5, V6~\cite{jainLiveCodeBenchHolisticContamination2024a} &  & New problems from coding contests \\
 & BigCodeBench-Hard \cite{zhuoBigCodeBenchBenchmarkingCode2024} &  & Practical and challenging programming problems \\
 & HumanEval+ \cite{liuYourCodeGenerated2023} &  & Formatted programming problems and its improved version \\ \hline
\end{tabular}
\caption{Information on benchmarks used in our study.}
\label{tab:appendix-benchmarks}
\end{table*}

\subsection{Analysis of LoRA Variants and Hyperparameters}
\input{ACL_2026/tabs/lora_ablation}

In Table~\ref{tab:lora_abaltion}, we compared the translation performance of Qwen3 on the FLORES-101 dataset under various LoRA variants and hyperparameter settings. We observed that the vanilla LoRA showed a clear advantage over other variants. Notably, low-rank LoRA achieved significantly better performance than high-rank configurations, which might be attributed to its stronger ability to mitigate catastrophic forgetting. Based on these observations, we adopted the LoRA tuning with rank = 8 in our main experiments.

%% file: ACL_2026/tabs/translation_flores_qwen3.tex
\begingroup
\renewcommand{\arraystretch}{1} 
\begin{table*}[!t]
\centering
\footnotesize
\resizebox{\linewidth}{!}{
\begin{tabular}{l|rr|rr|rr|rr|rr|rr|rr}

\toprule
& \multicolumn{2}{c|}{\textbf{x $\rightarrow$ en}} & \multicolumn{2}{c|}{\textbf{x $\rightarrow$ sw}} & \multicolumn{2}{c|}{\textbf{x $\rightarrow$ th}} & \multicolumn{2}{c|}{\textbf{x $\rightarrow$ bn}} & \multicolumn{2}{c|}{\textbf{x$\rightarrow$ zh}} & \multicolumn{2}{c|}{\textbf{x $\rightarrow$ ar}} & \multicolumn{2}{c}{\textbf{x $\rightarrow$ ko}} \\
& \textbf{spBLEU} & \textbf{xComet} & \textbf{spBLEU} & \textbf{xComet} & \textbf{spBLEU} & \textbf{xComet} & \textbf{spBLEU} & \textbf{xComet} & \textbf{spBLEU} & \textbf{xComet} & \textbf{spBLEU} & \textbf{xComet} & \textbf{spBLEU} & \textbf{xComet} \\

\midrule

Qwen3-8B & 35.24 & 89.89 & 3.49 & 12.52 & \textbf{29.85} & \textbf{75.47} & 14.14 & 59.91 & 26.88 & 81.65 & 21.73 & 75.18 & 16.20 & 76.48 \\
Qwen3-8B-FFT & 9.73 & 34.21 & 2.19 & 20.31 & 2.36 & 22.56 & 5.02 & 24.63 & 9.10 & 32.25 & 8.16 & 37.12 & 2.19 & 21.57 \\
Qwen3-8B-LoRA & 35.63 & 84.64 & 1.99 & 17.66 & 22.39 & 67.04 & 10.62 & 47.49 & 22.89 & 75.64 & 15.04 & 66.77 & 11.82 & 68.17 \\
\modelsmall & \textbf{38.02} & \textbf{91.35} & \textbf{18.60} & \textbf{50.99} & 27.84 & 73.17 & \textbf{19.39} & \textbf{70.67} & \textbf{26.95} & \textbf{82.15} & \textbf{24.00} & \textbf{77.50} & \textbf{18.08} & \textbf{80.54} \\

\midrule

Qwen3-14B & 36.92 & 91.98 & 5.87 & 15.33 & \textbf{32.40} & 80.12 & 17.50 & 68.09 & 28.71 & 83.95 & 24.01 & 79.75 & 18.77 & 82.19 \\
Qwen3-14B-FFT & \textbf{40.37} & 90.99 & 13.04 & 53.69 & 19.34 & 63.68 & 17.13 & 67.70 & 24.98 & 77.93 & 21.37 & 73.84 & 14.76 & 74.66 \\
Qwen3-14B-LoRA & 37.19 & 86.57 & 6.13 & 18.90 & 26.10 & 67.32 & 16.12 & 60.78 & 24.40 & 75.81 & 20.88 & 69.40 & 17.08 & 77.12 \\ 
\modellarge & 39.01 & \textbf{92.86} & \textbf{20.02} & \textbf{57.66} & 32.03 & \textbf{80.47} & \textbf{21.63} & \textbf{77.43} & \textbf{28.96} & \textbf{84.90} & \textbf{26.31} & \textbf{82.19} & \textbf{20.31} & \textbf{84.68} \\

\bottomrule

\toprule
& \multicolumn{2}{c|}{\textbf{en $\rightarrow$ x}} & \multicolumn{2}{c|}{\textbf{sw $\rightarrow$ x}} & \multicolumn{2}{c|}{\textbf{th $\rightarrow$ x}} & \multicolumn{2}{c|}{\textbf{bn $\rightarrow$ x}} & \multicolumn{2}{c|}{\textbf{zh $\rightarrow$ x}} & \multicolumn{2}{c|}{\textbf{ar $\rightarrow$ x}} & \multicolumn{2}{c}{\textbf{ko $\rightarrow$ x}} \\
& \textbf{spBLEU} & \textbf{xComet} & \textbf{spBLEU} & \textbf{xComet} & \textbf{spBLEU} & \textbf{xComet} & \textbf{spBLEU} & \textbf{xComet} & \textbf{spBLEU} & \textbf{xComet} & \textbf{spBLEU} & \textbf{xComet} & \textbf{spBLEU} & \textbf{xComet} \\

\midrule


Qwen3-8B & 28.86 & 80.27 & 13.61 & 32.32 & 19.68 & 72.64 & 19.38 & 66.51 & 20.82 & 78.90 & 22.85 & 72.27 & 20.23 & 71.79 \\
Qwen3-8B-FFT & 10.73 & 35.25 & 6.81 & 24.55 & 4.34 & 23.75 & 5.88 & 25.34 & 6.25 & 30.00 & 7.13 & 28.00 & 5.69 & 25.41 \\
Qwen3-8B-LoRA & 26.57 & 73.86 & 10.4 & 29.33 & 17.27 & 65.55 & 16.78 & 58.28 & 17.93 & 71.24 & 19.77 & 64.89 & 17.37 & 63.33 \\
\modelsmall & \textbf{32.82} & \textbf{85.52} & \textbf{21.41} & \textbf{55.06} & \textbf{22.68} & \textbf{77.36} & \textbf{22.47} & \textbf{71.75} & \textbf{23.31} & \textbf{83.13} & \textbf{25.46} & \textbf{75.63} & \textbf{22.73} & \textbf{75.77} \\

\midrule

Qwen3-14B & 31.78 & 84.54 & 18.40 & 43.75 & 22.35 & 78.09 & 22.47 & 72.32 & 23.02 & 83.04 & 25.28 & 76.98 & 22.90 & 77.07 \\
Qwen3-14B-FFT & 34.35 & 83.58 & 21.26 & \textbf{57.83} & 16.62 & 71.63 & 20.07 & 68.88 & 21.17 & 80.57 & 23.42 & 73.69 & 20.38 & 71.66 \\
Qwen3-14B-LoRA & 28.26 & 75.52 & 15.79 & 40.48 & 20.39 & 70.27 & 18.78 & 61.55 & 18.36 & 69.12 & 22.61 & 68.99 & 21.34 & 71.84 \\
\modellarge & \textbf{35.97} & \textbf{89.51} & \textbf{23.19} & 57.23 & \textbf{24.91} & \textbf{83.40} & \textbf{24.93} & \textbf{77.15} & \textbf{25.41} & \textbf{87.64} & \textbf{28.18} & \textbf{81.90} & \textbf{25.53} & \textbf{82.16}  \\

\bottomrule

\end{tabular}%
4}

\caption{Comparison of \modelseries with Qwen3, and between Full Fine-Tuning~(FFT) and LoRA on 17 languages from the FLORES-101 test set. In this table, “x” denotes translation into any of the other 16 languages, excluding the source and target languages in each translation direction.}
\label{tab:translation_flores_qwen3}
\end{table*}
\endgroup

%% file: ACL_2026/tabs/translation_flores.tex
\begingroup
\renewcommand{\arraystretch}{1} 
\begin{table*}[!ht]
\centering
\footnotesize
\resizebox{\linewidth}{!}{
\begin{tabular}{l|rc|rc|rc|rc|rc|rc|rc}

\toprule
& \multicolumn{2}{c|}{\textbf{x $\rightarrow$ en}} & \multicolumn{2}{c|}{\textbf{x $\rightarrow$ sw}} & \multicolumn{2}{c|}{\textbf{x $\rightarrow$ th}} & \multicolumn{2}{c|}{\textbf{x $\rightarrow$ bn}} & \multicolumn{2}{c|}{\textbf{x$\rightarrow$ zh}} & \multicolumn{2}{c|}{\textbf{x $\rightarrow$ ar}} & \multicolumn{2}{c}{\textbf{x $\rightarrow$ ko}} \\
& \textbf{spBLEU} & \textbf{xComet} & \textbf{spBLEU} & \textbf{xComet} & \textbf{spBLEU} & \textbf{xComet} & \textbf{spBLEU} & \textbf{xComet} & \textbf{spBLEU} & \textbf{xComet} & \textbf{spBLEU} & \textbf{xComet} & \textbf{spBLEU} & \textbf{xComet} \\

\midrule
\multicolumn{15}{c}{Super-Large Models} \\
\midrule

DeepSeek-R1-0528 & 1.85 & 14.73 & 0.57 & 13.98 & 5.01 & 28.10 & 0.33 & 14.92 & 19.84 & 67.62 & 1.53 & 16.87 & 1.35 & 16.68 \\
DeepSeek-V3-0324 & 33.59 & 79.36 & 16.60 & 47.84 & 24.37 & 60.62 & 16.5 & 60.33 & 29.55 & 86.96 & 9.27 & 40.02 & 14.40 & 59.19 \\
Kimi-K2 & 40.47 & 95.16 & 22.38 & 67.43 & 36.81 & 87.62 & 23.70 & 87.52 & 31.86 & 88.44 & 28.09 & 87.96 & 22.57 & 89.80 \\
Qwen3-235B-A22B & 41.19 & 94.80 & 16.01 & 41.90 & 35.81 & 87.51 & 24.37 & 82.90 & 31.39 & 87.98 & 27.92 & 87.11 & 21.59 & 88.85 \\

\midrule
\multicolumn{15}{c}{General LLMs} \\
\midrule

Gemma-3-12B-it & 36.82 & 86.44 & 0.63 & 13.46 & 2.00 & 19.13 & 0.90 & 18.61 & 22.36 & 69.97 & 1.20 & 17.63 & 7.41 & 39.49 \\
LLaMA3-8B & 0.38 & 14.98 & 2.38 & 17.27 & 1.45 & 18.1 & 6.46 & 33.12 & 1.79 & 21.52 & 2.78 & 18.07 & 0.13 & 11.11 \\
LLaMA3.1-8B & 30.96 & 79.59 & 5.83 & 32.84 & 16.55 & 50.92 & 13.80 & 59.67 & 17.35 & 64.60 & 9.42 & 40.48 & 11.02 & 60.44 \\
Qwen2.5-7B & 31.07 & 78.37 & 1.70 & 9.62 & 20.46 & 55.45 & 6.11 & 28.07 & 23.13 & 73.58 & 12.94 & 50.75 & 8.94 & 53.13 \\
Qwen2.5-14B & 25.89 & 64.86 & 1.58 & 13.00 & 8.68 & 37.33 & 6.50 & 31.96 & 22.75 & 69.12 & 7.98 & 33.72 & 6.64 & 38.73 \\
Qwen2.5-32B & 36.2 & 87.64 & 5.41 & 15.70 & 27.44 & 69.19 & 13.48 & 55.83 & 27.68 & 82.02 & 17.51 & 61.38 & 15.86 & 70.59 \\


\midrule

\multicolumn{15}{c}{Domain-Specialized LLMs} \\
\midrule

InternLM2-Math-7B & 21.67 & 64.71 & 0.29 & 24.34 & 0.19 & 28.25 & 0.08 & 21.61 & 4.75 & 45.37 & 1.19 & 24.13 & 0.24 & 19.35 \\
Deepseek-Math-7B & 21.94 & 71.32 & 0.11 & 23.71 & 3.26 & 25.90 & 0.98 & 22.08 & 12.14 & 59.94 & 0.78 & 22.60 & 2.35 & 37.64 \\
DeepSeek-Coder-V2-Lite & 29.61 & 79.00 & 1.15 & 11.12 & 14.06 & 34.95 & 5.04 & 27.76 & 21.90 & 70.26 & 11.82 & 43.43 & 9.66 & 49.13 \\

CodeLlama-7b & 20.65 & 63.32 & 0.52 & 24.76 & 0.34 & 34.47 & 0.08 & 25.97 & 0.63 & 47.69 & 0.07 & 24.45 & 0.40 & 32.94 \\
Qwen2.5-Math-7B & 2.29 & 17.48 & 0.09 & 14.04 & 0.08 & 20.61 & 0.02 & 20.2 & 0.32 & 19.81 & 0.03 & 22.46 & 0.13 & 20.23 \\
Qwen2.5-Coder-7B & 28.52 & 73.67 & 0.19 & 12.23 & 12.14 & 42.03 & 1.32 & 20.60 & 20.80 & 66.86 & 9.31 & 41.37 & 7.14 & 45.93 \\
Qwen2.5-Coder-14B & 30.16 & 75.40 & 0.67 & 13.52 & 7.95 & 36.35 & 1.06 & 24.91 & 21.13 & 73.30 & 5.65 & 30.52 & 3.63 & 34.00 \\
Qwen2.5-Coder-32B & 31.80 & 78.01 & 0.85 & 14.78 & 20.37 & 51.62 & 8.66 & 38.07 & 25.04 & 76.24 & 13.48 & 49.67 & 11.03 & 56.90 \\

\midrule
\multicolumn{15}{c}{Multilingual LLMs} \\
\midrule
TowerInstruct-7B-v0.1 & 29.26 & 72.56 & 0.71 & 37.34 & 0.48 & 53.90 & 0.25 & 59.70 & 17.02 & 62.16 & 0.48 & 58.62 & 13.4 & 62.28 \\
Hunyuan-MT-7B & 21.20 & 67.68 & 5.55 & 32.95 & 17.70 & 56.92 & 8.92 & 47.17 & 18.35 & 73.67 & 13.70 & 54.37 & 10.32 & 58.28 \\
Sailor2-8B-Chat & 0.52 & 17.81 & 0.67 & 17.09 & 24.12 & 60.84 & 2.42 & 20.67 & 16.69 & 60.53 & 5.60 & 31.41 & 4.46 & 37.03 \\
LLaMAX3-8B-Alpaca & 35.96 & 89.98 & 10.00 & 53.15 & 23.62 & 72.43 & 12.04 & 66.76 & 21.08 & 77.72 & 17.57 & 72.17 & 11.90 & 76.11 \\
Tower-Plus-9B & \textbf{40.12} & 91.74 & 2.45 & 20.80 & 18.71 & 53.76 & 2.47 & 58.16 & \textbf{30.37} & 82.96 & 9.66 & 48.73 & \textbf{22.36} & \textbf{85.53} \\
Aya-Expanse-8B & 33.13 & 79.28 & 1.49 & 8.91 & 6.42 & 19.81 & 4.94 & 25.08 & 23.53 & 70.67 & 23.77 & 70.21 & 17.71 & 70.53 \\
Aya-Expanse-32B & 39.72 & 88.63 & 2.60 & 16.53 & 15.16 & 40.65 & 11.93 & 53.76 & 27.93 & 80.70 & \textbf{28.63} & \textbf{81.70} & 21.71 & 82.79 \\

\midrule

\multicolumn{15}{c}{\modelseries} \\
\midrule

Qwen3-8B & 35.24 & 89.89 & 3.49 & 12.52 & 29.85 & 75.47 & 14.14 & 59.91 & 26.88 & 81.65 & 21.73 & 75.18 & 16.20 & 76.48 \\
Qwen3-8B-FT & 9.73 & 34.21 & 2.19 & 20.31 & 2.36 & 22.56 & 5.02 & 24.63 & 9.1 & 32.25 & 8.16 & 37.12 & 2.19 & 21.57 \\
Qwen3-8B-LoRA & 35.63 & 84.64 & 1.99 & 17.66 & 22.39 & 67.04 & 10.62 & 47.49 & 22.89 & 75.64 & 15.04 & 66.77 & 11.82 & 68.17 \\
\modelsmall & 38.02 & 91.35 & 18.60 & 50.99 & 27.84 & 73.17 & 19.39 & 70.67 & 26.95 & 82.15 & 24.00 & 77.50 & 18.08 & 80.54 \\
\midrule
Qwen3-14B & 36.92 & 91.98 & 5.87 & 15.33 & 32.40 & 80.12 & 17.50 & 68.09 & 28.71 & 83.95 & 24.01 & 79.75 & 18.77 & 82.19 \\
Qwen3-14B-FFT & 40.37 & 90.99 & 13.04 & 53.69 & 19.34 & 63.68 & 17.13 & 67.70 & 24.98 & 77.93 & 21.37 & 73.84 & 14.76 & 74.66 \\
Qwen3-14B-LoRA & 37.19 & 86.57 & 6.13 & 18.9 & 26.10 & 67.32 & 16.12 & 60.78 & 24.40 & 75.81 & 20.88 & 69.40 & 17.08 & 77.12 \\ 
\modellarge & 39.01 & 92.86 & 20.02 & 57.66 & 32.03 & 80.47 & 21.63 & 77.43 & 28.96 & 84.90 & 26.31 & 82.19 & 20.31 & 84.68 \\
\bottomrule

\toprule
& \multicolumn{2}{c|}{\textbf{en $\rightarrow$ x}} & \multicolumn{2}{c|}{\textbf{sw $\rightarrow$ x}} & \multicolumn{2}{c|}{\textbf{th $\rightarrow$ x}} & \multicolumn{2}{c|}{\textbf{bn $\rightarrow$ x}} & \multicolumn{2}{c|}{\textbf{zh $\rightarrow$ x}} & \multicolumn{2}{c|}{\textbf{ar $\rightarrow$ x}} & \multicolumn{2}{c}{\textbf{ko $\rightarrow$ x}} \\
& \textbf{spBLEU} & \textbf{xComet} & \textbf{spBLEU} & \textbf{xComet} & \textbf{spBLEU} & \textbf{xComet} & \textbf{spBLEU} & \textbf{xComet} & \textbf{spBLEU} & \textbf{xComet} & \textbf{spBLEU} & \textbf{xComet} & \textbf{spBLEU} & \textbf{xComet} \\

\midrule
\multicolumn{15}{c}{Super-Large Models} \\
\midrule

DeepSeek-R1-0528 & 2.91 & 20.74 & 1.18 & 14.28 & 1.90 & 18.81 & 1.72 & 18.12 & 10.85 & 48.22 & 1.92 & 15.68 & 2.40 & 20.09 \\
DeepSeek-V3-0324 & 30.83 & 74.27 & 15.84 & 43.38 & 20.20 & 64.89 & 21.47 & 65.26 & 25.52 & 84.73 & 24.31 & 66.01 & 23.44 & 72.22 \\
Kimi-K2 & 38.78 & 93.86 & 29.92 & 72.68 & 26.37 & 88.32 & 27.93 & 83.90 & 27.07 & 91.84 & 30.38 & 87.33 & 27.45 & 87.26 \\
Qwen3-235B-A22B & 36.84 & 91.37 & 27.34 & 66.69 & 26.54 & 85.61 & 27.19 & 81.47 & 26.42 & 88.48 & 29.68 & 84.62 & 26.80 & 84.10 \\

\midrule
\multicolumn{15}{c}{General LLMs} \\
\midrule

Gemma-3-12B-it & 7.23 & 44.02 & 4.42 & 20.07 & 7.06 & 30.65 & 6.86 & 31.32 & 8.99 & 39.85 & 9.60 & 34.06 & 9.33 & 36.39 \\
LLaMA3-8B & 18.77 & 57.18 & 0.07 & 11.41 & 2.38 & 22.78 & 0.45 & 15.49 & 1.42 & 19.48 & 0.38 & 12.24 & 1.87 & 21.38 \\
LLaMA3.1-8B & 26.97 & 81.09 & 8.96 & 29.03 & 15.06 & 65.86 & 16.62 & 59.50 & 10.74 & 51.34 & 10.61 & 42.32 & 16.73 & 63.44 \\

Qwen2.5-7B & 19.39 & 59.90 & 5.03 & 15.02 & 14.20 & 54.89 & 10.74 & 41.27 & 12.91 & 55.85 & 14.37 & 48.59 & 12.75 & 47.20 \\
Qwen2.5-14B & 6.90 & 31.92 & 4.26 & 19.95 & 10.34 & 41.12 & 8.55 & 38.32 & 10.66 & 43.37 & 10.04 & 36.82 & 7.78 & 33.41 \\
Qwen2.5-32B & 27.58 & 74.29 & 14.34 & 35.24 & 18.94 & 66.27 & 17.94 & 59.77 & 17.26 & 64.52 & 21.65 & 65.39 & 19.00 & 63.87 \\


\midrule
\multicolumn{15}{c}{Domain-Specialized LLMs} \\
 \midrule

CodeLlama-7B & 4.81 & 71.10 & 0.15 & 11.70 & 0.70 & 15.41 & 0.21 & 12.45 & 1.82 & 47.68 & 1.04 & 17.83 & 1.61 & 33.68 \\
InternLM2-Math-7B & 7.26 & 40.40 & 1.13 & 13.76 & 1.64 & 21.55 & 0.83 & 19.17 & 4.25 & 49.02 & 2.71 & 39.27 & 2.48 & 35.73 \\

DeepSeek-Math-7B & 12.97 & 53.45 & 0.79 & 15.72 & 3.36 & 38.27 & 2.49 & 30.42 & 6.17 & 55.35 & 3.78 & 31.81 & 5.52 & 50.14 \\
DeepSeek-Coder-V2-Lite & 21.82 & 63.34 & 5.82 & 17.29 & 10.66 & 45.73 & 9.95 & 36.05 & 13.66 & 58.76 & 13.57 & 48.63 & 12.48 & 47.68 \\

Qwen2.5-Math-7B & 0.44 & 38.89 & 0.03 & 14.01 & 0.16 & 14.33 & 0.04 & 14.80 & 0.36 & 28.90 & 0.11 & 11.74 & 0.19 & 12.55 \\
Qwen2.5-Coder-7B & 17.53 & 59.59 & 3.96 & 13.81 & 10.13 & 43.65 & 7.33 & 28.57 & 11.26 & 56.78 & 10.70 & 41.13 & 10.73 & 44.55 \\
Qwen2.5-Coder-14B & 19.19 & 61.28 & 2.78 & 15.87 & 6.86 & 36.47 & 6.65 & 31.96 & 13.13 & 59.08 & 8.66 & 39.29 & 7.30 & 38.95 \\
Qwen2.5-Coder-32B & 20.94 & 62.04 & 8.00 & 23.64 & 11.47 & 47.78 & 11.20 & 42.60 & 15.61 & 61.88 & 13.46 & 47.14 & 13.19 & 49.97 \\

\midrule
\multicolumn{15}{c}{Multilingual LLMs} \\
 \midrule

TowerInstruct-7B-v0.1 & 18.26 & 67.80 & 2.67 & 17.33 & 3.82 & 29.22 & 1.78 & 21.56 & 10.89 & 69.26 & 7.95 & 39.09 & 11.26 & 70.69 \\
Hunyuan-MT-7B & 28.43 & 87.96 & 14.12 & 39.86 & 7.53 & 36.37 & 4.60 & 28.71 & 20.37 & 83.94 & 15.72 & 55.10 & 14.31 & 51.93 \\
Sailor2-8B-Chat & 16.03 & 54.11 & 1.76 & 14.08 & 4.30 & 33.86 & 5.72 & 30.13 & 3.83 & 28.77 & 7.29 & 34.26 & 7.18 & 36.11 \\
LLaMAX3-8B-Alpaca & 26.62 & 83.22 & 20.23 & 59.06 & 16.03 & 74.88 & 17.20 & 68.49 & 16.51 & 80.33 & 18.37 & 73.77 & 17.39 & 72.23 \\
Tower-Plus-9B & 28.83 & 79.85 & 18.38 & 46.19 & 19.02 & 70.29 & 18.64 & 62.92 & 19.39 & 75.55 & 21.72 & 69.28 & 20.68 & 71.93 \\
Aya-Expanse-8B & 25.75 & 68.36 & 7.90 & 16.43 & 11.39 & 40.78 & 11.29 & 36.77 & 17.85 & 65.45 & 20.21 & 60.86 & 18.41 & 60.74 \\
Aya-Expanse-32B & 30.21 & 78.30 & 16.72 & 38.82 & 18.25 & 64.11 & 19.37 & 62.04 & 21.26 & 75.21 & 24.71 & 71.16 & 22.07 & 70.84 \\

\midrule
\multicolumn{15}{c}{\modelseries} \\
\midrule
Qwen3-8B & 28.86 & 80.27 & 13.61 & 32.32 & 19.68 & 72.64 & 19.38 & 66.51 & 20.82 & 78.90 & 22.85 & 72.27 & 20.23 & 71.79 \\
Qwen3-8B-FFT & 10.73 & 35.25 & 6.81 & 24.55 & 4.34 & 23.75 & 5.88 & 25.34 & 6.25 & 30.0 & 7.13 & 28.00 & 5.69 & 25.41 \\
Qwen3-8B-LoRA & 26.57 & 73.86 & 10.40 & 29.33 & 17.27 & 65.55 & 16.78 & 58.28 & 17.93 & 71.24 & 19.77 & 64.89 & 17.37 & 63.33 \\
\modelsmall & 32.82 & 85.52 & 21.41 & 55.06 & 22.68 & 77.36 & 22.47 & 71.75 & 23.31 & 83.13 & 25.46 & 75.63 & 22.73 & 75.77 \\
\midrule
Qwen3-14B & 31.78 & 84.54 & 18.40 & 43.75 & 22.35 & 78.09 & 22.47 & 72.32 & 23.02 & 83.04 & 25.28 & 76.98 & 22.90 & 77.07 \\
Qwen3-14B-FFT & 34.35 & 83.58 & 21.26 & 57.83 & 16.62 & 71.63 & 20.07 & 68.88 & 21.17 & 80.57 & 23.42 & 73.69 & 20.38 & 71.66 \\
Qwen3-14B-LoRA & 28.26 & 75.52 & 15.79 & 40.48 & 20.39 & 70.27 & 18.78 & 61.55 & 18.36 & 69.12 & 22.61 & 68.99 & 21.34 & 71.84 \\
\modellarge & 35.97 & 89.51 & 23.19 & 57.23 & 24.91 & 83.40 & 24.93 & 77.15 & 25.41 & 87.64 & 28.18 & 81.90 & 25.53 & 82.16  \\

\bottomrule

\end{tabular}%
}

\caption{Translation results on 17 languages from FLORES-101 test set. In this table, “x” denotes translation into any of the other 16 languages, excluding the source and target languages in each translation direction.}
\label{tab:translation_flores}
\end{table*}
\endgroup

%% file: ACL_2026/tabs/understanding_results.tex
\begin{table*}[!ht]
\centering
\resizebox{\textwidth}{!}{%
\begin{tabular}{@{}l|c|c|c|c|c|c|c|c|c@{}}
\toprule
 \textbf{Models} & \textbf{HumanEval$+$} & \textbf{XNLI} & \textbf{MGSM} & \textbf{xIFEval} & \textbf{XStoryCloze} & \textbf{MathQA} & \textbf{XCOPA} & \textbf{XGPQA} & \textbf{XWinograd} \\
\midrule 




Aya-Expanse-8B & 40.24 & 45.53 & 14.51 & 40.46 & 64.80 & 37.69 & 56.36 & 20.85 & 74.67 \\
CodeLlama-7B & 31.71 & 40.69 & 1.78 & 30.51 & 56.45 & 28.71 & 54.58 & 14.40 & 76.42 \\
DeepSeek-Coder-V2-Lite & 74.39 & 42.30 & 30.62 & 43.41 & 62.69 & 45.16 & 59.27 & 22.86 & 80.24 \\
LLaMAX3-8B-Alpaca & 24.39 & 45.33 & 9.78 & 36.74 & 61.84 & 34.17 & 63.85 & 21.78 & 74.80 \\
Llama-3-8B & 54.27 & 42.14 & 55.38 & 65.32 & 59.50 & 27.24 & 60.89 & 27.25 & 71.09 \\
Llama-3.1-8B & 61.59 & 44.83 & 28.36 & 45.46 & 64.47 & 39.20 & 60.89 & 19.80 & 81.70 \\
Qwen2.5-14B & 73.17 & 39.64 & 39.78 & 58.35 & 66.72 & 48.48 & 65.33 & 32.41 & 82.72 \\
Qwen2.5-32B & 84.15 & 40.16 & 74.95 & 82.67 & 65.91 & 38.22 & 65.98 & 38.12 & 71.66 \\
Qwen2.5-7B & 75.00 & 39.84 & 34.80 & 57.64 & 63.25 & 40.27 & 62.84 & 26.19 & 80.58 \\
Qwen2.5-Coder-7B & 85.37 & 42.95 & 52.25 & 61.01 & 58.58 & 37.15 & 58.96 & 23.83 & 70.60 \\
Sailor2-8B-Chat & 39.63 & 38.30 & 34.84 & 43.01 & 63.56 & 40.64 & 24.11  & 24.11 & 81.64 \\
Tower-Plus-9B & 0.00 & 42.54 & 64.00 & 75.52 & 61.65 & 33.13 & 60.62 & 27.36 & 72.02 \\
TowerInstruct-7B-v0.1 & 19.51 & 40.45 & 3.75 & 30.19 & 59.11 & 29.25 & 56.98 & 15.77 & 78.40 \\
DeepSeek-Math-7B & 51.83 & 42.02 & 26.95 & 33.33 & 58.56 & 37.96 & 56.67 & 20.25 & 76.26 \\
Gemma-3-12B-IT & 0.00 & 33.33 & 0.04 & 25.19 & 52.81 & 20.57 & 50.00 & 0.00 & 51.63 \\
InternLM2-Math-7B & 28.05 & 38.60 & 39.24 & 35.63 & 55.68 & 26.20 & 55.45 & 18.12 & 61.50 \\
\midrule
Qwen3-8B & 76.83 & 42.44 & 44.87 & 81.68 & 58.08 & 32.80 & 60.40 & 35.85 & 63.79 \\
\modelsmall & 76.22 & 44.79 & 50.36 & 80.49 & 59.24 & 32.93 & 61.44 & 34.26 & 67.95 \\
\midrule
Qwen3-14B & 85.37 & 43.05 & 52.18 & 85.64 & 58.73 & 34.71 & 61.87 & 41.43 & 60.58 \\
\modellarge & 85.98 & 44.77 & 50.22 & 85.55 & 61.14 & 35.88 & 63.60 &	40.10 &	64.87 \\
\bottomrule
\end{tabular}%
}
\caption{Comparison of \modelseries and existing LLMs on reasoning and multilingual tasks.}
\label{tab:general_task}
\end{table*}

%% file: ACL_2026/tabs/lora_ablation.tex
\begin{table*}[!ht]
\centering
\scriptsize
\resizebox{\linewidth}{!}{%
\begin{tabular}{@{}c|c|ll|ll|c|c|ll|ll@{}}
\toprule
\multirow{3}{*}{\textbf{setting}} & \multicolumn{1}{c|}{\multirow{3}{*}{\textbf{rank}}} & \multicolumn{2}{c|}{\textbf{language$\rightarrow$x}} & \multicolumn{2}{c|}{\textbf{x$\rightarrow$language}} & \multirow{3}{*}{\textbf{setting}} & \multicolumn{1}{c|}{\multirow{3}{*}{\textbf{rank}}} & \multicolumn{2}{c|}{\textbf{language$\rightarrow$x}} & \multicolumn{2}{c}{\textbf{x$\rightarrow$language}} \\
&  & \textbf{spBLEU} & \textbf{xComet} & \textbf{spBLEU} & \textbf{xComet} &  &  & \textbf{spBLEU} & \textbf{xComet} & \textbf{spBLEU} & \textbf{xComet} \\ 
\midrule
\multicolumn{6}{c|}{Qwen3-8B} & \multicolumn{6}{c}{Qwen3-14B} \\
\midrule
\multirow{5}{*}{LoRA} & 8 & 18.01 & 60.93 & 17.83 & 61.49 & \multirow{5}{*}{LoRA} & 8 & 20.79 & 65.40 & 21.13 & 65.13 \\
 & 16 & 16.83 & 57.76 & 16.46 & 57.67 &  & 16 & 13.45 & 47.44 & 14.64 & 49.51 \\
 & 32 & 15.87 & 56.05 & 16.05 & 56.35 &  & 32 & 10.91 & 40.80 & 13.01 & 44.22 \\
 & 64 & 15.31 & 54.90 & 14.28 & 54.03 &  & 64 & 6.75 & 32.50 & 8.15 & 35.30 \\
 & 128 & 13.95 & 51.42 & 11.93 & 47.05 &  & 128 & 7.55 & 37.87 & 8.04 & 40.55 \\
 \midrule
\multirow{5}{*}{rsLoRA} & 8 & 15.58 & 55.44 & 16.35 & 57.56 & \multirow{5}{*}{rsLoRA} & 8 & 8.29 & 35.53 & 10.64 & 39.48 \\
 & 16 & 13.64 & 51.81 & 11.32 & 47.82 &  & 16 & 5.52 & 30.83 & 6.55 & 33.21 \\
 & 32 & 12.51 & 48.02 & 10.41 & 41.80 &  & 32 & 8.66 & 44.68 & 9.86 & 46.39 \\
 & 64 & 14.60 & 59.01 & 13.18 & 56.66 &  & 64 & 9.96 & 48.65 & 10.59 & 49.81 \\
 & 128 & 13.17 & 52.54 & 11.26 & 49.56 &  & 128 & 9.59 & 43.24 & 9.51 & 44.66 \\
 \midrule
\multirow{5}{*}{DLoRA} & 8 & 17.49 & 60.16 & 17.20 & 61.06 & \multirow{5}{*}{DLoRA} & 8 & 12.90 & 47.87 & 15.85 & 53.31 \\
 & 16 & 16.62 & 57.77 & 16.28 & 58.14 &  & 16 & 9.68 & 39.09 & 11.75 & 42.75 \\
 & 32 & 15.53 & 55.49 & 15.78 & 56.29 &  & 32 & 8.32 & 35.26 & 10.45 & 38.73 \\
 & 64 & 15.52 & 55.93 & 14.82 & 55.69 &  & 64 & 5.62 & 30.57 & 6.85 & 32.50 \\
 & 128 & 14.07 & 52.06 & 12.43 & 48.10 &  & 128 & 8.93 & 43.40 & 10.23 & 44.50 \\
 \bottomrule
\end{tabular}%
}
\caption{Translation performance of Qwen3 under different LoRA variants and hyperparameter settings.}
\label{tab:lora_abaltion}
\end{table*}